\documentclass[final,3p,times]{elsarticle}

\usepackage{amssymb}
\biboptions{comma,sort&compress}
\usepackage{amsmath}
\usepackage{xcolor}
\usepackage{hyperref}       
\usepackage{url}            
\usepackage{booktabs}       
\usepackage{nicefrac}       
\usepackage{microtype}      

\usepackage{caption}
\usepackage{multicol}
\usepackage{graphicx}
\usepackage{subfig}
\usepackage{tabularx}
\usepackage{latexsym}
\usepackage{amstext,amsfonts,amssymb,amsmath}
\usepackage{todonotes}
\usepackage[version=3]{mhchem}
\usepackage{epstopdf}
\AppendGraphicsExtensions{.tif}
\usepackage{color}
\usepackage{bm}
\usepackage{algorithm}
\usepackage{algpseudocode}
\newcommand\Algphase[1]{%
\vspace*{-.7\baselineskip}\Statex\hspace*{\dimexpr-\algorithmicindent-2pt\relax}\rule{\textwidth}{0.4pt}%
\Statex\hspace*{-\algorithmicindent}\textbf{#1}%
\vspace*{-.7\baselineskip}\Statex\hspace*{\dimexpr-\algorithmicindent-2pt\relax}\rule{\textwidth}{0.4pt}%
}
\usepackage{multirow}
\usepackage{tikz}
\usepackage{breqn}
\usepackage{etoolbox}

\usepackage{tikz}
\usepackage{breqn}
\usepackage{etoolbox}

\newrobustcmd*{\myVtriangle}[2]{\tikz{\filldraw[draw=#1,fill=#2] (0cm,0.2cm) --
(0.2cm,0.2cm) -- (0.1cm,0cm) -- (0cm,0.2cm);}}

\newrobustcmd*{\mythickVtriangle}[2]{\tikz{\filldraw[line width=0.3mm,draw=#1,fill=#2] (0cm,0.2cm) --
(0.2cm,0.2cm) -- (0.1cm,0cm) -- (0cm,0.2cm);}}

\newrobustcmd*{\mythickErrorVtriangle}[2]{\tikz{\filldraw[line width=0.3mm,draw=#1,fill=#2] (-0.05cm,0.05cm) --
(0.05cm,0.05cm) -- (0cm,-0.05cm) -- (-0.05cm,0.05cm);  \draw[draw=#1] (0.0cm, -0.12cm) -- (0.0cm, 0.12cm) ; \draw[draw=#1] (-0.06cm, 0.12cm) -- (0.06cm, 0.12cm); \draw[draw=#1] (-0.06cm, -0.12cm) -- (0.06cm, -0.12cm)    }}

\newrobustcmd*{\mytriangle}[2]{\tikz{\filldraw[draw=#1,fill=#2] (0.0cm,0.0cm) --
(0.2cm,0cm) -- (0.1cm,0.2cm) -- (0cm,0cm);}}

\newrobustcmd*{\mysquare}[2]{\tikz{\draw[draw=#1,fill=#2] (0cm,0cm)
rectangle (0.2cm,0.2cm)}}

\newrobustcmd*{\mythicktriangle}[2]{\tikz{\filldraw[line width=0.3mm,draw=#1,fill=#2] (0.0cm,0cm) --
(0.2cm,0cm) -- (0.1cm,0.2cm) -- (0.0cm,0cm);}}

\newrobustcmd*{\mythicksquare}[2]{\tikz{\draw[line width=0.3mm,draw=#1,fill=#2] (0cm,0cm)
rectangle (0.2cm,0.2cm)}}

\newrobustcmd*{\mybarredtriangle}[2]{\tikz{\draw[draw=#1,fill=#2] (0,0) --
(0.2cm,0) -- (0.1cm,0.2cm) -- (0cm,0cm); \draw[draw=#1] (-0.1cm, 0.07cm) -- (0.3cm, 0.07cm)}}

\newrobustcmd*{\mythickbarredtriangle}[2]{\tikz{\draw[line width=0.3mm,draw=#1,fill=#2] (0,0) --
(0.2cm,0) -- (0.1cm,0.2cm) -- (0cm,0cm); \draw[draw=#1] (-0.1cm, 0.07cm) -- (0.3cm, 0.07cm)}}

\newrobustcmd*{\mybarredsquare}[2]{\tikz{\draw[draw=#1,fill=#2] (0,0)
rectangle (0.2cm,0.2cm); \draw[draw=#1] (-0.1cm, 0.1cm) -- (0.3cm, 0.1cm)}}

\newrobustcmd*{\mythickbarredsquare}[2]{\tikz{\draw[line width=0.3mm,draw=#1,fill=#2] (0,0)
rectangle (0.2cm,0.2cm); \draw[draw=#1] (-0.1cm, 0.1cm) -- (0.3cm, 0.1cm)}}

\newrobustcmd*{\mybarredcircle}[2]{\tikz{\draw[draw=#1,fill=#2] (0,0)
circle (0.1cm); \draw[draw=#1] (-0.2cm, 0.0cm) -- (0.2cm, 0.0cm)}}

\newrobustcmd*{\mythickbarredcircle}[2]{\tikz{\draw[line width=0.3mm,draw=#1,fill=#2] (0,0)
circle (0.1cm); \draw[draw=#1] (-0.2cm, 0.0cm) -- (0.2cm, 0.0cm)}}

\newrobustcmd*{\mythickErrorcircle}[2]{\tikz{\draw[line width=0.3mm,draw=#1,fill=#2] (0,0)
circle (0.06cm); \draw[draw=#1] (0.0cm, -0.12cm) -- (0.0cm, 0.12cm) ;   \draw[draw=#1] (-0.06cm, 0.12cm) -- (0.06cm, 0.12cm); \draw[draw=#1] (-0.06cm, -0.12cm) -- (0.06cm, -0.12cm)    }}

\newrobustcmd*{\mydashedline}[1]{\tikz{\draw[draw=#1] (-0.2cm, 0.2cm) -- (-0.1cm, 0.2cm); \draw[draw=#1] (-0.0cm, 0.2cm) -- (0.1cm, 0.2cm)}}

\newrobustcmd*{\mythickcross}[1]{\tikz{\draw[line width=0.3mm,draw=#1] (0,0) --
(0.2cm,0); \draw[line width=0.3mm,draw=#1] (0.1cm,-0.1cm) -- (0.1cm,0.1cm);}}

\newrobustcmd*{\mybarredcross}[1]{\tikz{\draw[line width=0.3mm,draw=#1] (0,0) --
(0.2cm,0); \draw[line width=0.3mm,draw=#1] (0.1cm,-0.1cm) -- (0.1cm,0.1cm); \draw[draw=#1] (-0.1cm,0) -- (0.3cm,0);}}

\newrobustcmd*{\myline}[1]{\tikz{\draw[draw=#1] (-0.15cm, 0.1cm) -- (0.15cm, 0.1cm);\draw[line width=0.3mm,draw=#1] (-0.0cm, 0.0cm);}}

\newrobustcmd*{\mythickline}[1]{\tikz{\draw[line width=0.3mm,draw=#1] (-0.15cm, 0.1cm) -- (0.15cm, 0.1cm);\draw[line width=0.3mm,draw=#1] (-0.0cm, 0.0cm);}}

\newrobustcmd*{\mythickdashedline}[1]{\tikz{\draw[line width=0.3mm,draw=#1] (-0.2, 0.1cm) -- (-0.1cm, 0.1cm); \draw[line width=0.3mm,draw=#1] (-0.0cm, 0.1cm) -- (0.1cm, 0.1cm); \draw[line width=0.3mm,draw=#1] (-0.0cm, 0.0cm);}}

\newrobustcmd*{\mythickdasheddottedline}[1]{\tikz{\draw[line width=0.3mm,draw=#1] (-0.22, 0.1cm) -- (-0.13cm, 0.1cm); \draw[line width=0.3mm,draw=#1] (-0.085cm, 0.1cm) -- (-0.055cm, 0.1cm); \draw[line width=0.3mm,draw=#1] (-0.01cm, 0.1cm) -- (0.08cm, 0.1cm); \draw[line width=0.3mm,draw=#1] (-0.0cm, 0.0cm);}}

\newrobustcmd*{\mycircle}[2]{\tikz{\draw[draw=#1,fill=#2] (0,0)
circle (0.1cm);}}

\newrobustcmd*{\mythickcircle}[2]{\tikz{\draw[line width=0.3mm,draw=#1,fill=#2] (0,0)
circle (0.1cm);}}

\newrobustcmd*{\mydot}[1]{\tikz{\draw[line width=0.3mm,draw=#1] (0,0)
circle (0.025cm);}}

\journal{Journal}

\begin{document}

\begin{frontmatter}

\title{Uniform-in-Phase-Space Data Selection with Iterative Normalizing Flows}
\author[fir]{Malik Hassanaly\corref{cor1}}
\ead{malik.hassanaly@nrel.gov}
\author[fir]{Bruce A. Perry}
\author[sec,fir]{Michael E. Mueller}
\author[fir]{Shashank Yellapantula}
\address[fir]{Computational Science Center, National Renewable Energy Laboratory, 80401 Golden, CO, USA}
\address[sec]{Mechanical and Aerospace Engineering, Princeton University, 08544 Princeton, NJ, USA}
\cortext[cor1]{Corresponding author:}

\begin{abstract}
Improvements in computational and experimental capabilities are rapidly increasing the amount of scientific data that is routinely generated. In applications that are constrained by memory and computational intensity, excessively large datasets may hinder scientific discovery, making data reduction a critical component of data-driven methods. Datasets are growing in two directions: the number of data points and their dimensionality. Whereas dimension reduction typically aims at describing each data sample on lower-dimensional space, the focus here is on reducing the number of data points. A strategy is proposed to select data points such that they uniformly span the phase-space of the data.
The algorithm proposed relies on estimating the probability map of the data and using it to construct an acceptance probability. An iterative method is used to accurately estimate the probability of the rare data points when only a small subset of the dataset is used to construct the probability map. Instead of binning the phase-space to estimate the probability map, its functional form is approximated with a normalizing flow. Therefore, the method naturally extends to high-dimensional datasets. The proposed framework is demonstrated as a viable pathway to enable data-efficient machine learning when abundant data is available.

\end{abstract}

\begin{graphicalabstract}
\begin{figure}[h!]
\centering
\includegraphics[width=0.99\linewidth]{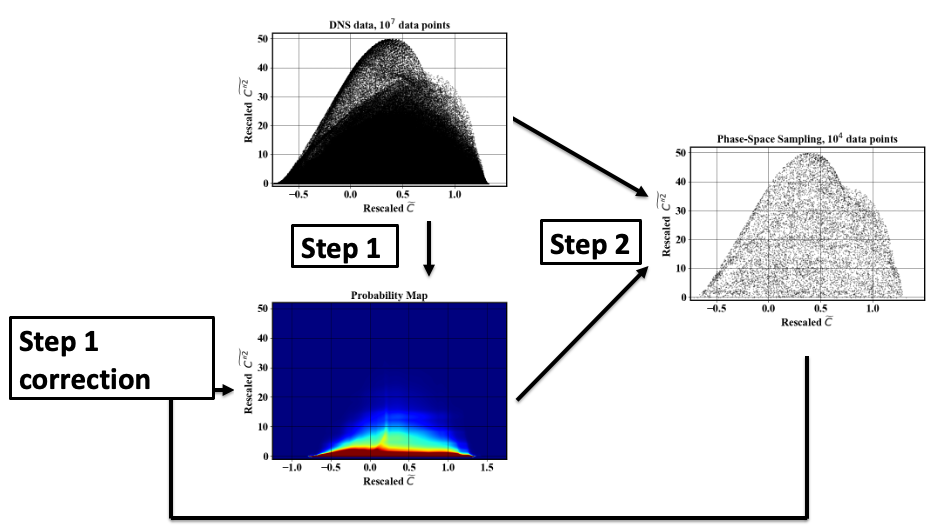}
\caption{Graphical illustration of the data selection method. Starting from a large dataset (top left scatter plot), a reduced uniform-in-phase-space dataset is constructed (right scatter plot). The reduction method relies on the probability map which is iteratively computed (Step 1 and Step 1 correction). An acceptance probability constructed with the probability map is used to accept or reject each data point (Step 2).}
\label{fig:Ill}
\end{figure}
\end{graphicalabstract}


\end{frontmatter}

\tableofcontents
\section{Introduction}
\label{sec:intro}

Advances in high-performance computing (HPC) and experimental diagnostics have led to the generation of ever-growing datasets. Scientific applications have since successfully embraced the era of Big Data \citep{brunton2020machine,duraisamy2019turbulence,baker2019workshop} but are now struggling to handle very large datasets. For example, numerical simulations of combustion or fusion can generate hundreds of petabytes per day, thereby heavily stressing storage and network resources \citep{klasky2021data}. Likewise, experimental facilities generate measurements with ever-higher spatial and temporal resolution \citep{stohr2019time,barwey2019experimental}. In this context, data reduction is becoming a critical part of scientific machine learning (ML) \citep{peterka2020priority}.   


A scientific dataset can be thought of as an element of $\mathbb{R}^{N\times D}$, where $N$ is the number of data points and $D$ is the dimension of a data point. Data reduction can first be achieved by recognizing that, although the data is $D$-dimensional, it may in effect lie on a low-dimensional manifold of size $d \ll D$. With an appropriate $d$-dimensional decomposition, the full dataset will be well approximated. This approach is commonly termed \textit{dimension reduction} and has received extensive attention \citep{brunton2020machine,hinton2006reducing,oseledets2011tensor,sirovich1987turbulence}. The dimension reduction strategy is also at the heart of projection-based reduced-order models \citep{gouasmi2017priori,akram2022approximate}.

Another approach consists of discarding some data points from the dataset, i.e., representing the dataset with $n \ll N$ data points, which is commonly termed \textit{instance selection} (IS) \citep{jankowski2004comparison}, \textit{data-sampling} \citep{biswas2018situ,woodring2011situ}, or \textit{data-pruning} \citep{saseendran2019impact}. Other data compression methods such as tensor-decomposition do not distinguish between data dimensions and adopt a holistic approach \citep{ballard2020tuckermpi}. Although there are well-established concepts that ensure that minimal information is lost during \textit{dimension reduction}, data reconstruction may still be necessary when disseminating data or performing scientific tasks. For example, dimension reduction techniques alone do not ease the visualization of large datasets. In contrast, IS does not require reconstruction and naturally eases scientific tasks. Note that if data reconstruction is not needed, IS can be used in conjunction with dimension reduction for more aggressive data reduction. In this paper, a new method for IS is proposed for large $N$ and large $D$.


IS can be an attractive method for improving the accuracy of a model that uses a given dataset. For example, in the case of classification tasks, if one class is over-represented compared to another (class imbalance), it may be advantageous to remove some data points from the majority class. Several techniques have been successful, including random removal of majority class samples \citep{leevy2018survey}. More advanced techniques have proposed removing noisy data points (instance editing)~\citep{tomek1976two,angelova2005pruning,wilson1972asymptotic}, removing data points located away from the decision boundary (instance condensation)~\citep{hart1968condensed}, or using a hybrid of both approaches \citep{batista2004study}. Alternatively, the pruned dataset can be constructed iteratively by testing its classification accuracy \citep{garcia2009evolutionary,lopez2014addressing,skalak1994prototype}. These approaches are particularly well suited for nearest-neighbor classification, which exhibits degrading performance with increasing training dataset size \citep{wilson2000reduction,garcia2012prototype}. In natural language processing, IS has also been shown to be useful for improving the accuracy of predictive models \citep{mudrakarta2018training}. Despite the existence of a wide variety of approaches, the aforementioned pruning techniques often do not scale linearly with the number of data points in the original dataset and may not be appropriate to address Big Data in scientific applications \citep{jankowski2004comparison,triguero2016evolutionary}. 


In some situations, IS may simply be unavoidable --- for instance, when so much data is being generated that some needs be discarded due to storage constraints. In particular, although parallel visualization tools are available, they may be cumbersome and intensive for computing and network resources. Besides, for very large datasets, input/output time may become the bottleneck of numerical simulations \citep{woodring2011situ}. To address this issue, researchers have proposed various methods for spatially downselecting data points while maintaining the relevant visual features. ~\citet{woodring2011situ} proposed a stratified sampling method for astronomical particle data such that the downselected dataset maintains the original spatial distribution of particles. A variant of adaptive mesh refinement for data was also proposed to guide downsampling, where the refinement criterion is user-defined \citep{nouanesengsy2014adr}. For the visualization of scatter plots, a perceptual loss was proposed to assess whether the scatter plot of downselected data was consistent with the original dataset \citep{park2016visualization}. Methods similar to the one proposed here have been proposed in the past~\citep{rapp2019void,biswas2018situ}, but either do not scale with the number of dimensions ($D$) or instances ($N$). \citet{rapp2019void} proposed a method to display a multidimensional scatter plot by transforming the distribution into an arbitrary one. However, this method may not scale well with the dimensionality of the data, as it requires a neighbor search for the density estimation. In addition, the authors show that the method can become unreasonably slow for very large datasets. In the same vein, \citet{biswas2018situ} proposed performing spatial downsampling by achieving uniform sampling of scalar quantities. They first estimate the probability density function (PDF) of the scalar values and then use it to downselect data points. The method was later extended to include the scalar gradients to ensure visual smoothness \citep{biswas2020probabilistic}. This technique requires binning the phase-space to construct the PDF, which poses two main issues. First, the choice of the number of bins can affect the quality of the results \citep{biswas2020probabilistic}. Second, binning high-dimensional spaces can become intractable in terms of computational intensity and memory, which makes extension to uniform sampling in higher scalar dimensions nontrivial \citep{dutta2019multivariate}. 

Besides visualization, IS may also be needed for other scientific tasks in power and memory-constrained environments, which may become ubiquitous with the rise of the Internet of Things ~\citep{raman2019emerging}. In a practical edge computing case, data may simply be randomly discarded to reduce memory and power requirements. This is especially true for streaming data, as newer data is typically retained, whereas old data is discarded to adapt to drift in the observations \citep{ramirez2017survey,hulten2001mining,krawczyk2015one}. In such cases, the IS method adopted is critical for the performance of the device trained with the streaming data \citep{du2014detecting}. Finally, memory requirements that make IS unavoidable do not solely exist for edge devices, but can also occur in supercomputing environments. Consider the example of ML applied to turbulence modeling. Typically, high-fidelity data --- such as direct numerical simulation (DNS) --- is used to train ML-based closure models. For this application, each location in physical space is a new data point. Given that a single DNS snapshot nowadays comprises $\mathcal{O}(10^9)$ data points, multiple snapshots can easily amount to $\mathcal{O}(10^{11})$ training data points. With this number of data points, training time may become unreasonable, and data may need to be discarded. Data is often randomly discarded, but more elaborate methods have recently been proposed, such as clustering the data and randomly selecting data from each cluster \citep{lloyd1982least,nguyen2021machine,yellapantula2021deep}. This technique enables the efficient elimination of redundant data points from quiescent or otherwise unimportant regions when the phenomenon of interest is intermittent, effectively attempting an approximate phase-space sampling like the one proposed in the present work. However, IS via clustering is sensitive to the choice of the number of clusters, and there is no systematic way of deciding on an appropriate number of clusters \citep{barwey2019experimental,hassanaly2021data,barwey2020data}. Furthermore, in practice, clustering tends to underrepresent rare data points (see Appendix~\ref{app:visualInspection}). In separate studies, researchers have proposed clustering the dataset and training a separate model for each cluster. This technique is beneficial not only for a priori \citep{barwey2021data} but also a posteriori analyses \citep{nguyen2021machine}, suggesting that a proper downsampling technique could help with the deployment of ML models.

In this work, a novel method for IS in large and high-dimensional datasets is proposed. Redundant samples are pruned based on their similarity in the original $D$-dimensional space. This is achieved by first estimating a probability map of the original dataset and then using it to uniformly sample phase-space. The proposed solution estimates the probability map with a normalizing flow, which has been shown to perform well in high-dimensional systems. In addition, an iterative method is proposed for the probability map estimation, thereby efficiently and accurately estimating the probability map without processing the full dataset. As such, the computational cost may be drastically reduced compared to methods that process the full dataset. The proposed algorithm is described at length in Sec.~\ref{sec:method}. In Sec.~\ref{sec:results}, the uniformity of the pruned dataset is assessed and the scaling of the computational cost with respect to the number of data points and their dimension is evaluated. The effect of the uniform sampling on ML tasks is also compared to other sampling methods in Sec.~\ref{sec:dataEfficient}. Section~\ref{sec:conclusions} provides conclusions and perspectives for extending the proposed approach.

\section{Instance selection method}
\label{sec:method}


\subsection{Problem statement and justification}

Consider a dataset $\mathcal{X} \in \mathbb{R}^{N \times D}$, where $N$ is the number of instances (or data points) and $D$ is the number of dimensions (or features) of the dataset. The objective is to create a smaller dataset $\mathcal{Y} \in \mathbb{R}^{n \times D}$, where $n \ll N$, and where every instance of $\mathcal{Y}$ is included in $\mathcal{X}$. In other words, $\mathcal{Y}$ does not contain new data points. In addition, during the downselection process, the redundant samples should be pruned first. Redundancy is defined as proximity in the $D$-dimensional phase-space. The objective is that all regions of phase-space should have the same density of data points, i.e., the distribution of $\mathcal{Y}$ should be as uniform as possible in phase-space.  

Compared to a random sampling approach, uniform sampling of the phase-space emphasizes rare events in the reduced dataset. In contrast, random downsampling will almost always discard rare events from the dataset. The motivation behind preserving rare data points is twofold. First, scientific discovery often stems from rare events. For example, state transitions may occur rarely, but observing them is critical for prediction, causality inference, and control \citep{barwey2019experimental,hassanaly2021classification}. If a dataset is reduced without appropriate treatment, scientific interpretation may be hindered. Second, to ensure that data-driven models are robust, they need to be exposed to a dataset that is as diverse as possible. Although the average training error may only be mildly affected, the model may still incur large errors in rare parts of phase-space. In some applications, accurate prediction of rare events is crucial when the model is deployed. For example, in non-premixed rotating detonation engines, detonations are heavily influenced by triple points and fuel mixing ahead of the detonation \citep{sato2021mixing,barwey2021data,rankin2017chemiluminescence}, which are very localized phenomena. For other combustion applications, the lack of data in some parts of phase-space has even required the addition of synthetic data \citep{wan2020chemistry}. The proposed sampling objective has also been noted to maximize the information content of the subsampled dataset~\citep{biswas2018situ}.

\subsection{Method overview}
\label{sec:overview}

To perform the downsampling task, the first step is to estimate the PDF of the dataset in phase-space. Once this probability map has been constructed, it is used to construct an acceptance probability. The acceptance probability determines whether a given data point should be selected or not. In practice, the acceptance probability is used as follows. For each data point, a random number is drawn from the uniform distribution $\mathcal{U}(0,1)$. If this number falls below the acceptance probability, the point is integrated into the downsampled dataset. The acceptance probability varies for each data point and depends on the likelihood of observing such data. In regions of high data density, the acceptance probability is low, but in regions of low data density (rare regions), the acceptance probability is high. The link between data PDF and the appropriate acceptance probability is illustrated hereafter for a simple example.

Consider two events, $A$ and $B$, observed respectively $a$ and $b$ times, where $b < a$. Suppose that one wants to select $c$ samples such that $c \leq 2b < a + b$. In that case, the acceptance probability $s(A)$ of every event $A$ (or $s(B)$ of every event $B$) should be such that the expected number of event $A$ (or $B$) in the downsampled dataset is $c/2$. In other words, $s(A) \times a = c/2$ (or $s(B) \times b = c/2$). Therefore, the acceptance probability should be proportional to the inverse of the PDF, with the proportionality factor controlled by the desired size of the downsampled set. The specific data points selected depend on the random number drawn from $\mathcal{U}(0,1)$ for each of the data points. This is a source of randomness in the proposed algorithm that is studied in Sec.~\ref{sec:dataEfficient}. 

\begin{figure}[h!]
    \centering
    \includegraphics[width=0.30\textwidth]{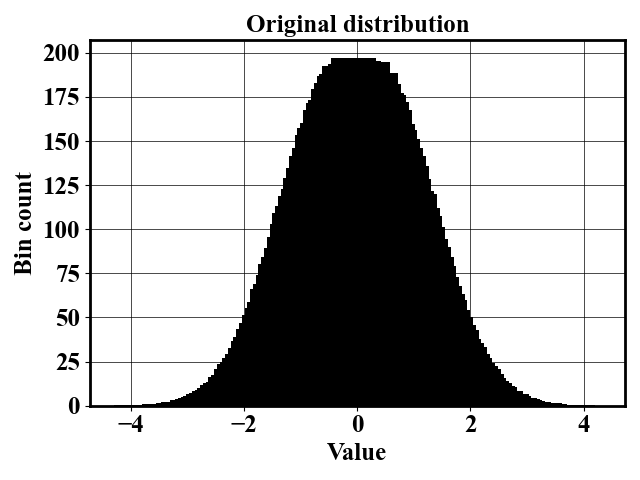}
    \includegraphics[width=0.30\textwidth]{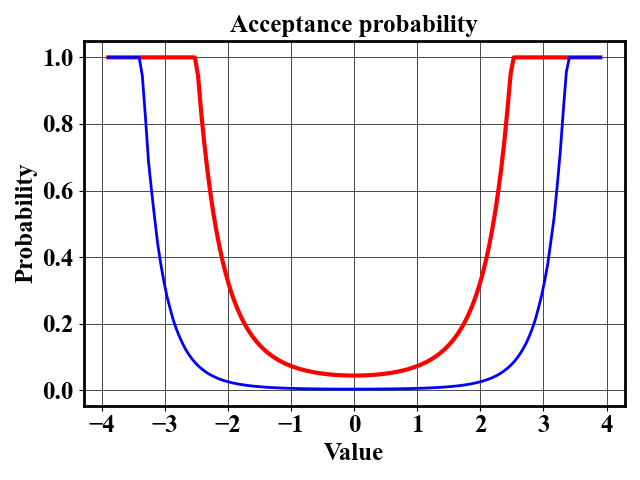}
    \includegraphics[width=0.30\textwidth]{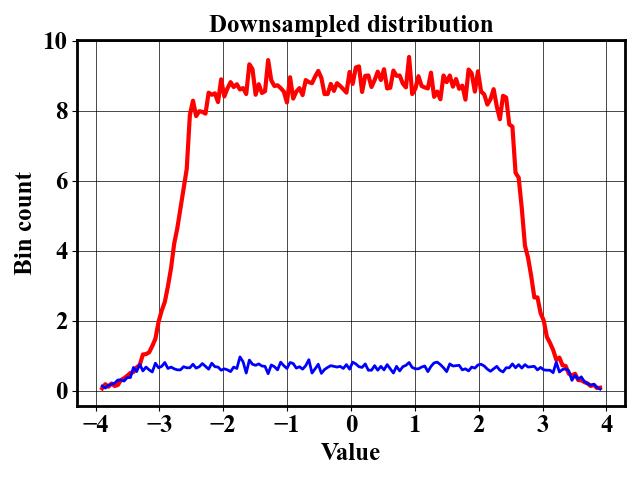}
    \caption{Illustration of the proposed method for a canonical example. Left: histogram of the original distribution. Middle: acceptance probability plotted against the random variable value when downsampling to $n=100$ data points (\mythickline{blue}) and $n=1,000$ data points (\mythickline{red}). Right: distribution of the downsampled dataset when downsampling to $n=100$ data points (\mythickline{blue}) and $n=1,000$ data points (\mythickline{red})}
    \label{fig:normal}
\end{figure}

In the event where $2b< c < a+b$, all $B$ events should be selected. Although the resulting downsampled dataset is imbalanced, it is as close as possible to a uniform distribution. To accommodate for this case, the acceptance probability $s(x)$ for the data point $x$ can be simply constructed as

\begin{equation}
    \label{eq:probClip}
    s(x) = min(\frac{\alpha}{p(x)},1), 
\end{equation}

\noindent where $x$ is an observation, $p(x)$ is the PDF of the dataset evaluated in $x$, and $\alpha$ is a proportionality factor constant for the whole dataset constructed such that the expected number of selected data points equals $c$, i.e.,
\begin{equation}
    \label{eq:alphaEq}
    \sum_{i=1}^N s(x_i) = c,
\end{equation}

\noindent where $x_i$ is the $i^{th}$ observation and $c$ is the desired number of samples. Using a constant proportionality parameter $\alpha$ for the full dataset ensures that the relative importance of every event is maintained. The method extends to continuous distributions, as only the local PDF values are needed to construct the acceptance probability. Fig.~\ref{fig:normal} shows an application of the method for selecting 100 and 1,000 data points from 10,000 samples distributed according to a standard normal distribution $\mathcal{N}(0,1)$. Here, the exact PDF is used to construct the acceptance probability. Then, the acceptance probability is adjusted to satisfy Eq.~\ref{eq:alphaEq}. The downsampled distributions are obtained by repeating the sampling algorithm 100 times. As can be seen in the figure, the distribution of samples approaches a uniform distribution in the high-probability region. For a large number of samples, the tail of the PDF for the downselected dataset is underrepresented due to the low data availability. In those regions, the acceptance probability is maximal and all the samples are selected. Note that to obtain exactly 1,000 samples (compared to 100 samples), a larger part of the dataset needs to be selected with maximal acceptance probability. Despite achieving uniform sampling where possible, the method cannot ensure a perfect balance between all events in the dataset. Because the acceptance probability is clipped at unity (i.e., the method does not create new data), rare events can be underrepresented in the reduced dataset if $c$ is large. The larger $c$ is, the more imbalance can be expected in the dataset. Addressing the imbalance in the reduced dataset would require generating new data points and is out of the scope of this paper. In the case where uniform sampling needs to be achieved, reducing the number of data points in the final dataset will help extending the volume of phase-space over which data is uniformly distributed. This effect can be observed in Fig.~\ref{fig:normal} (right) where using a lower $n$ value extends the range over which uniform sampling is achieved.

\subsection{Practical implementation}

Practical implementation of the method requires overcoming several numerical challenges, which are outlined below. In particular, because the objective is to tackle the issues associated with very large datasets, scalability with the number of instances and dimensions is necessary. A predictor procedure and an predictor-corrector procedure are proposed.

\subsubsection{Probability map estimation}

The main hurdle to overcome is the estimation of the PDF of the dataset in phase-space, which is ultimately used to compute the acceptance probability. Different solutions have been proposed, including kernel methods \citep{rapp2019void} and binning \citep{biswas2018situ}. Although the binning strategy is scalable with respect to the number of data points, the quality of the estimation may be dependent on the number of bins used \citep{biswas2020probabilistic}. {Kernel methods offer a more continuous description of the PDF but tend to generate spurious modes in high-dimension \cite{wang2019nonparametric}. In addition, the PDF may be sensitive to the choice of the smoothing parameter \cite{tabak2013family}. Finally, as local density value requires knowledge of the relative locations of all data points, specific treatments are needed such as using compact kernels, but in this case, the neighborhood of every point needs to be identified, which can be problematic with many dimensions \citep{rapp2019void}.} 

To estimate the density of the dataset, it is proposed to use normalizing flows, which are primarily applied to generative modeling \citep{dinh2014nice,dinh2016density,durkan2019neural,muller2019neural,kingma2016improving,papamakarios2017masked} and are increasingly being used for scientific applications \citep{bothmann2020exploring,verheyen2020phase}. Given a dataset, generative modeling attempts to create new data points that were previously unseen but are distributed like the original dataset. Some notable applications of generative modeling include semi-supervised learning \citep{salimans2016improved}, sampling of high-dimensional PDFs \citep{hassanaly2022adversarial,hassanaly2022ganisp}, and closure modeling \citep{bode2021using}. Normalizing flows are types of neural networks that generate new data by simultaneously learning the PDF of the dataset and generating data points with maximum likelihood (or minimum negative log-likelihood). The likelihood of the data is computed by using a transformation $G$ to map from a known random variable $z$ onto a random variable $x$ that is distributed like the dataset. The likelihood of the generated data can be computed as

\begin{equation}
    \label{eq:probTransf}
    p(x) = p(z) ~ \text{det} ( \frac{dG^{-1}}{dx} ).
\end{equation}

The peculiarity of normalizing flows lies in their attempt to directly evaluate the right-hand side of Eq.~\ref{eq:probTransf} allowing to evaluate $p(x)$. Normalizing flows build a mapping $G$ such that $\text{det} (\frac{dG^{-1}}{dx} )$ is tractable, even in high dimensions. In general, evaluating the determinant in high dimensions scales at least quadratically with the matrix dimension. Normalizing flows adopt structures that make the determinant of the Jacobian of the inverse of $G$ easy to compute. For instance, if the Jacobian is triangular, the cost of the determinant calculation scales linearly with data dimension ~\citep{dinh2014nice,dinh2016density}. The attention given to scaling with data dimension has allowed normalizing flows to be used for image generation, which shows that they can learn PDFs in a ${\sim}10^3$-dimensional phase-space. Recent types of normalizing flows typically attempt to find a balance between neural network expressiveness and the tractability of the right-hand side of Eq.~\ref{eq:probTransf}~\citep{dinh2014nice,dinh2016density,durkan2019neural}. Among them, neural spline flows are chosen here, as they have been shown to capture complex multimodal distributions with a small number of trainable parameters~\citep{durkan2019neural}. Importantly, the method proposed here may be easily adapted to future other density estimation methods that may tackle high-dimensions or handle large numbers of data points more efficiently.

\subsubsection{Predictor algorithm}
\label{sec:singlepassAlgo}

Instead of using a normalizing flow to generate new samples, it is proposed here to leverage its density estimation capabilities to construct an appropriate acceptance probability. Given $N$ instances of the dataset, training a normalizing flow requires processing all $N$ instances at every epoch. This approach is computationally expensive and therefore unacceptable, especially when the ultimate goal is to downselect data to efficiently develop an ML-based model. It is proposed to train a normalizing flow on a random subset of the data, thereby greatly accelerating the PDF estimation process. The algorithm is summarized in Algo.~\ref{algo:base}.

\begin{algorithm}
\caption{Predictor instance selection of $n$ data points}\label{algo:base}
\begin{algorithmic}[1]
\State Shuffle dataset $\mathcal{X} \in \mathbb{R}^{N \times d}$.
\State Create a working dataset $\xi \in \mathbb{R}^{M \times d}$, where $M \ll N$, by randomly probing $\mathcal{X}$.
\Algphase{Compute acceptance probability $s$}
\State Train a normalizing flow with working dataset to learn the probability map $p$. (Step 1)
\For{$x_i \in \mathcal{X}$ }
    \State Evaluate $p(x_i)$ and acceptance probability $S(x_i) = 1/p(x_i)$. (Step 2a)
\EndFor
\Algphase{Select $n$ data points (Step 2b)}
\State Find $\alpha$ such that $\sum_{i=1}^N min(max(\alpha s(x_i),0),1) = n$.
\State numberSelectedDataPoints = 0
\For{$i=1,N$}
    \State Draw a random number $r$ from the uniform distribution $\mathcal{U}(0,1)$.
    \If{$r<min(max(\alpha s(x_i),0),1)$ and numberSelectedDataPoints $<n$}
        \State Append the data point to the selected instances.
        \State numberSelectedDataPoints += 1.
    \EndIf
\EndFor
\end{algorithmic}
\end{algorithm}

\begin{figure}[h!]
    \centering
    \includegraphics[width=0.47\textwidth]{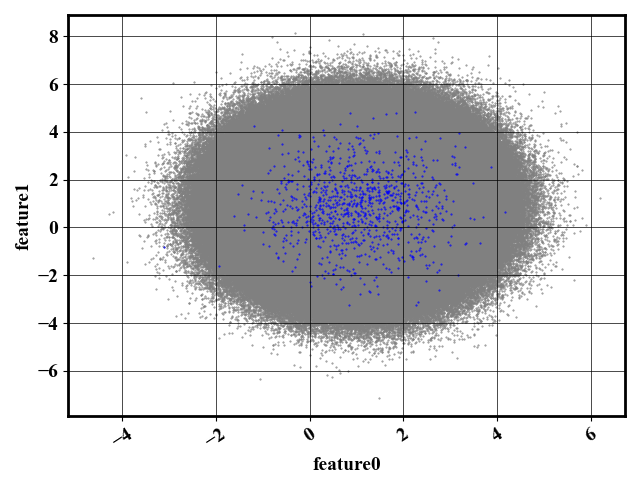}
    \includegraphics[width=0.47\textwidth]{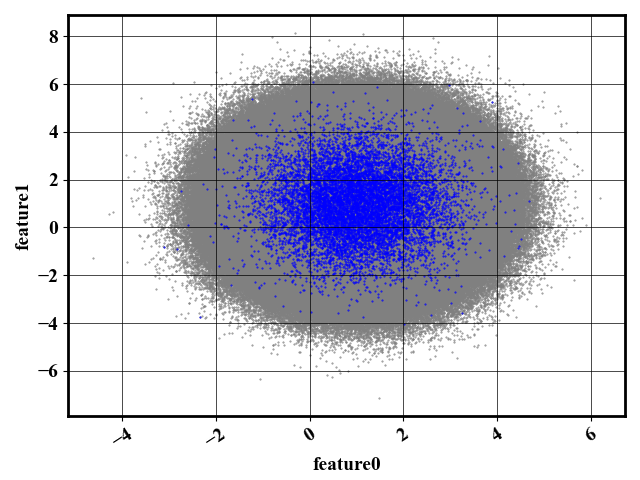}
    \includegraphics[width=0.47\textwidth]{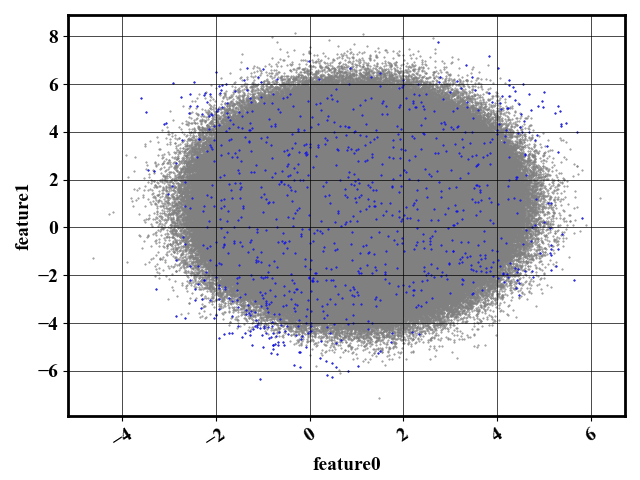}
    \includegraphics[width=0.47\textwidth]{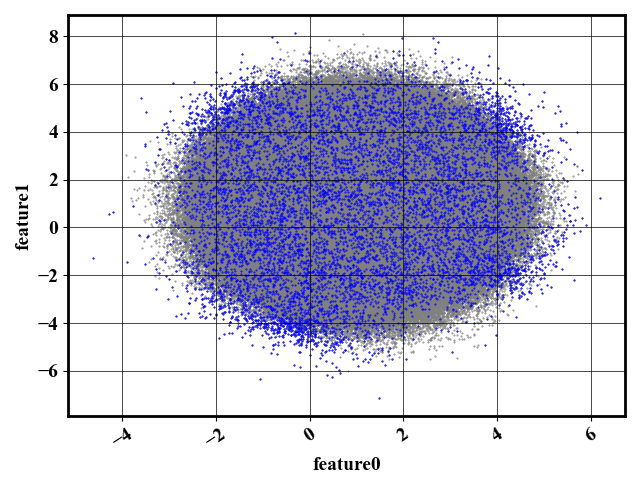}
    \caption{Illustration of the results obtained with Algo.~\ref{algo:base}. The full dataset (\mydot{gray}) is overlaid with the reduced dataset (\mydot{blue}). Top left: random selection of $n=10^3$ data points. Bottom left: result of Algo.~\ref{algo:base} with $n=10^3$. Top right: random selection of $n=10^4$ data points. Bottom right: result of Algo.~\ref{algo:base} with $n=10^4$}
    \label{fig:basemethod}
\end{figure}

The outcome of the base algorithm is illustrated in Fig.~\ref{fig:basemethod}. The dataset considered here is a bivariate normal distribution of mean $[1,1]$ and diagonal covariance matrix $diag([1,2])$ that contains $10^6$ samples. For computational efficiency, the probability map is learned with $M=10^5$ data points that are randomly selected (sensitivity to the choice of $M$ is assessed in Sec.~\ref{sec:sensitivityM}). A scatter plot of the randomly selected data is shown in Fig.~\ref{fig:basemethod} for $10^3$ and $10^4$ selected data points. Compared to a random sampling method (left), Algo.~\ref{algo:base} provides better coverage of phase-space, no matter the value of $n$. However, a non-uniform distribution of the data points can be observed, especially in low-probability regions.  This problem is addressed in the next section.

\subsubsection{Accommodating for rare observations: the predictor-corrector algorithm}
\label{sec:iterativeAlgo}

\begin{figure}[h!]
    \centering
    \includegraphics[width=0.47\textwidth]{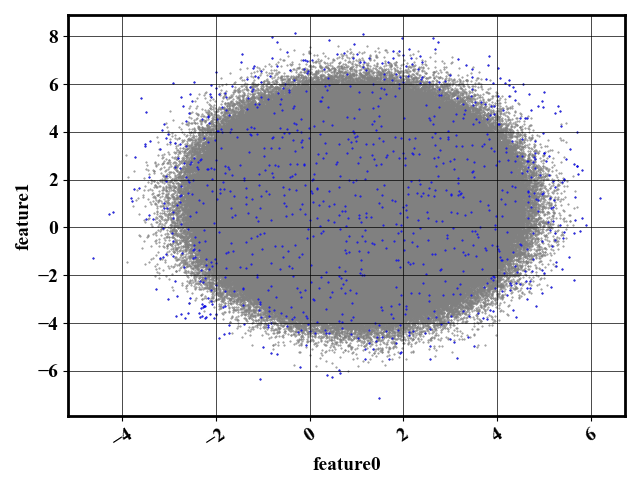}
    \includegraphics[width=0.47\textwidth]{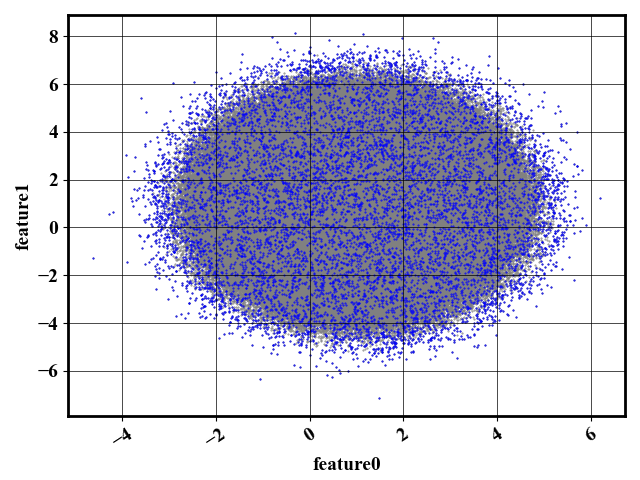}
    \caption{Left: result of Algo.~\ref{algo:iterative} with $n=10^3$. Right: result of Algo.~\ref{algo:iterative} with $n=10^4$}
    \label{fig:iterativemethod}
\end{figure}

The downside of accelerating the normalizing flow calculation by using a random subsample of the dataset when learning its PDF is that the probability of rare events is not well approximated, simply because too few rare events are shown to the normalizing flow. However, one can recognize that the wrongly downsampled data shown in Fig.~\ref{fig:basemethod} is an image of the error in the density estimation obtained from the normalizing flow. By computing the density of the downsampled data itself, one can apply a correction to the original density estimate such that, if Algo.~\ref{algo:base} is used again, the resulting data distribution is close to being uniform. The underlying idea that motivates the algorithm is that is it easier to learn a correction to the density than the full density itself. Similar methods have been developed in the context of multifidelity methods \citep{de2022bi}. The predictor-corrector is summarized in Algo.~\ref{algo:iterative}. In the predictor-corrector algorithm, the normalizing flow is trained $nFlowIter$ times, where $nFlowIter \in \mathbb{N^*}-\{1\}$. The procedure is used on the same dataset as in Sec.~\ref{sec:singlepassAlgo}, and the results are shown in Fig.~\ref{fig:iterativemethod}. 

\begin{algorithm}
\caption{Predictor-corrector instance selection of $n$ data points}\label{algo:iterative}
\begin{algorithmic}[1]
\State Shuffle dataset $\mathcal{X} \in \mathbb{R}^{N \times d}$.
\State Create a working dataset $\xi \in \mathbb{R}^{M \times d}$, where $M \ll N$, by randomly probing $\mathcal{X}$.
\For{flowIter$=1,$nFlowIter}
    \State Compute acceptance probability $s_{flowIter}$ (line 3--6 of Algo.~\ref{algo:base}).
    \If{flowIter==1}
        \State Select $M$ data points (line 7--15 of Algo.~\ref{algo:base}).
        \State Replace working dataset with downselected dataset
    \Else
        \State $s_{flowIter} = s_{flowIter} \times s_{flowIter-1}$.
        \If{flowIter==nFlowIter}
            \State Select $n$ data points (line 7--15 of Algo.~\ref{algo:base}).
        \Else
            \State Select $M$ data points (line 7--15 of Algo.~\ref{algo:base}).
            \State Replace working dataset with downselected dataset
        \EndIf
    \EndIf
\EndFor
\end{algorithmic}
\end{algorithm}


\begin{figure}[h!]
    \centering
    \includegraphics[width=0.47\textwidth]{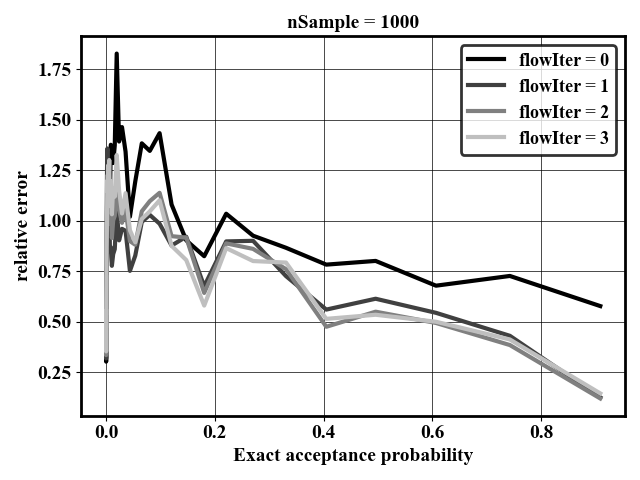}
    \includegraphics[width=0.47\textwidth]{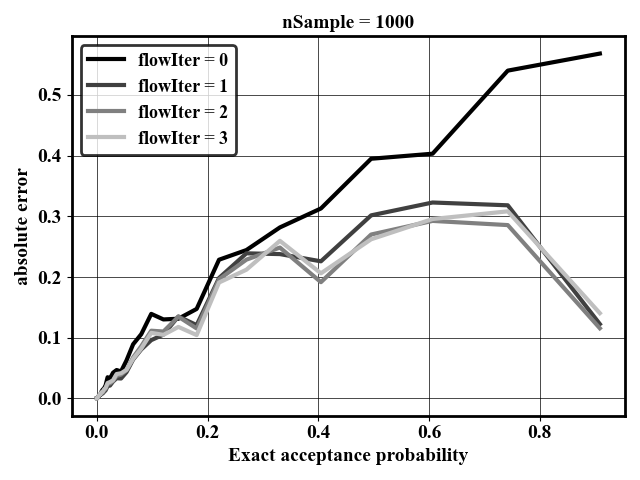}
    \caption{Effect of iterations on accuracy of acceptance probability. Left: conditional mean of relative error in the acceptance probability plotted against the exact acceptance probability. Right: conditional mean of absolute error in the acceptance probability plotted against the exact acceptance probability}
    \label{fig:conditionalError}
\end{figure}

For the specific case considered here, because the exact PDF is known, the true acceptance probability can be computed and compared to the sampling probability obtained at every iteration of the algorithm. Figure~\ref{fig:conditionalError} shows the relative error in the acceptance probability after different numbers of iterations. It can be seen that, as expected, iterating decreases errors for rare points (high acceptance probabilities). In addition, although errors decrease after the first iteration, they remain stagnant. Error stagnation can be explained by the fact that iterating is not expected to decrease the modeling errors of the normalizing flow --- only the statistical errors associated with estimating rare event probabilities.

\begin{figure}[h!]
    \centering
    \includegraphics[width=0.47\textwidth]{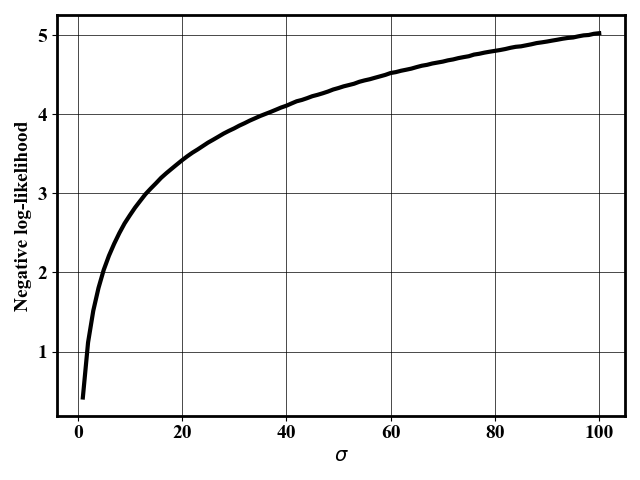}
    \includegraphics[width=0.47\textwidth]{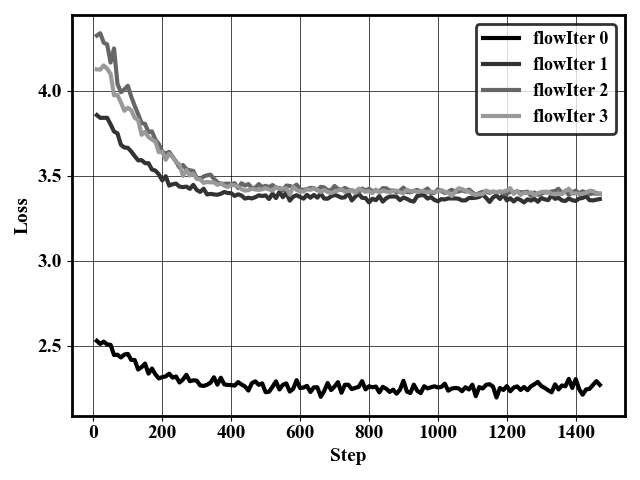}
    \caption{Left: negative log-likelihood against standard deviation for a standard normal distribution. Right: history of the negative log-likelihood loss of the normalizing flow trained at each flow iteration}
    \label{fig:normalNLL}
\end{figure}

Finally, some important properties for the predictor-corrector algorithm are described hereafter. First, after every iteration, the loss function (the negative log-likelihood) of the converged normalizing flow should increase. This can be understood by computing the exact negative log-likelihood of a normal distribution with increasing standard deviation (left-hand panel of Fig.~\ref{fig:normalNLL}). As the standard deviation increases, the negative log-likelihood increases as well. Compared to random sampling, the first iteration of the method should spread the data distribution. If that is the case, the second iteration of the predictor-corrector algorithm should result in a loss (negative log-likelihood) larger than at the first iteration. This behavior can be observed for the bivariate normal case (right-hand panel of Fig.~\ref{fig:normalNLL}).

Second, the PDF corrected with the iterative procedure will not approach the true PDF, but rather the PDF needed to obtain a uniform sampling of the data. This principle can be understood with a simple bimodal case such as the one in Sec.~\ref{sec:overview}. In the case where $c>2b$, the downsampled distribution simply cannot be exactly uniform. While iterating, even with an exact estimation of the data density at the first iteration, the density correction at the next iteration will be nonzero, because a uniform distribution of data is not possible (see discussion at the end of Sec.~\ref{sec:overview}). However, the rescaling factor $\alpha$ will ensure that the density estimate corrections do not affect the downsampled distribution, which is what matters here.

\subsubsection{Sampling probability adjustment}

When computing the proportionality factor $\alpha$ that satisfies Eq.~\ref{eq:alphaEq}, it is necessary to perform summations and probability clipping (Eq.~\ref{eq:probClip}) over the entire dataset. When there are a large number of instances ($N$), the cost of this step may approach the cost of density estimation. However, $\alpha$ can be computed with only a subset of all the acceptance probabilities. To see this, recall that the normalizing constant $\alpha$ must satisfy $\sum_{i=1}^{N} \alpha s(x_i) = n$, where $x_i$ is the $i^{th}$ data point. Since $\alpha$ does not depend on the data point considered, the problem simplifies to estimating $\frac{1}{N} \sum_{i=1}^{N} s(x_i)$. Since the data points are shuffled, the choice of the first $N'$ data points where $N'\ll N$ can be considered as $N'$ random draw of the random variable $s(x)$. Using the central limit theorem, $\frac{1}{N'}\sum_{i=1}^{N'} s(x_i)$ is a sample of a normal distribution centered on $\frac{1}{N}\sum_{i=1}^{N} s(x_i)$ and with a standard deviation $\sigma/\sqrt{N'}$ where $\sigma$ is the standard deviation of the random variable $s(x)$. The probability of making a large error in the estimation of $\frac{1}{N}\sum_{i=1}^{N} s(x_i)$ depends on $\sigma$ and the number of samples $N'$. The decision on the value of $N'$ is therefore case dependent. Note however that since $\forall i, 0<s(x_i)<1$ , $\sigma<1/2$ following Popoviciu's inequality. Therefore, the number of required $N'$ to achieve a given accuracy can be computed a priori and does not increase with $N$.
Using $N^{\prime} \ll N$ acceptance probability values, the constraint used to compute $\alpha$ becomes
\begin{equation}
    \frac{N}{N^{\prime}}\sum_{i=1}^{N^{\prime}} s(x_i) = c.
\end{equation}
With this approximation, and the ability to compute the probability map with a subset of the full dataset, the only task of Algo.~\ref{algo:iterative} that scales with the number of data points is the evaluation of the probability with the normalizing flow. In Sec.~\ref{sec:computationalcost}, a parallel strategy is described to further reduce the cost of the procedure.

During the selection process -- when the acceptance probability is used to decide which data point to keep in the downsampled dataset -- the order of the data points can matter. For instance, if the most rare data points are seen first, then those data points will be downselected before a sufficient amount of high-probability points are observed. To avoid this issue, it is crucial to randomly shuffle the dataset (first step of Algo.~\ref{algo:base} and Algo.~\ref{algo:iterative}). Finally, since the selection process is probabilistic, the probability adjustment only guarantees that $n$ data points is the expected number of data points to be selected. In practice, the number of datapoints selected may vary. If more than $n$ data points are selected, the first $n$ data points can be kept. If less than $n$ data points are selected, then another pass through the data can be done to complete the dataset.  

\section{Numerical experiments}
\label{sec:results}

In this section, the proposed method is tested on a real scientific dataset. The sensitivities of the proposed algorithm to the main hyperparameters (number of flow iterations, size of dataset chosen to train the normalizing flow) are assessed. The performance of the algorithm in high dimensions is evaluated. The computational cost of the method and the efficiency of parallelization strategies are studied.

\subsection{Dataset}
\label{sec:dataset}

\begin{figure}[h!]
    \centering
    \includegraphics[width=0.47\textwidth]{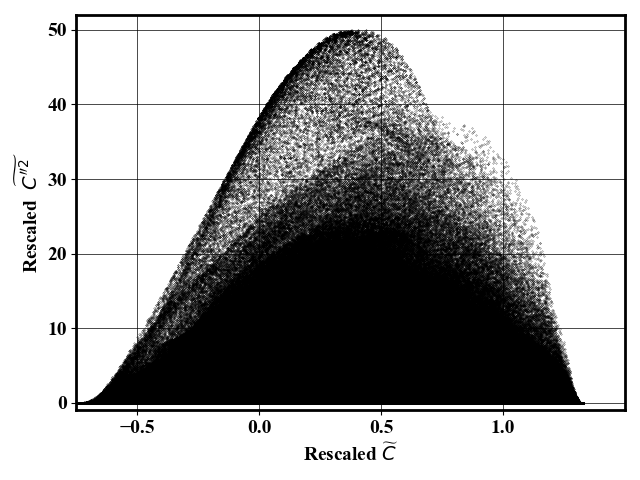}
    \caption{Scatter plot of the full dataset with $D=2$}
    \label{fig:dataset}
\end{figure}

The dataset considered here is the one used by \citet{yellapantula2021deep} in a turbulent combustion modeling application. In particular, the data was used to learn a model for the filtered dissipation rate of a reaction progress variable $C$ in large eddy simulations of turbulent premixed combustion systems. Dissipation rate is an unclosed term in the transport equation of the progress variable variance $\widetilde{C^{'' 2}}$, which is useful to estimate the filtered progress variable source term. The dataset was constructed by aggregating the data from multiple DNS datasets under different operating conditions. For computational tractability, the DNS was reduced with a stratified sampling approach, but it still exhibits a multimodal aspect, as can be seen in Fig.~\ref{fig:dataset}. The features of the dataset are $\{ \widetilde{C}, \widetilde{C^{''2}}, \widetilde{\chi}_{res}, \alpha, \beta, \gamma \}$, where $\widetilde{C}$ is a filtered progress variable, $\widetilde{C^{''2}}$ is the subfilter variance of the progress variable, $\widetilde{\chi}_{res} = 2 D_C |\nabla \widetilde{C}|^2$ is the resolved dissipation rate, $D_C$ is the progress variable diffusivity, and  $\alpha$, $\beta$, and $\gamma$ are the principal rates of strain. Additional details about the definitions of the dimensions are available in \citet{yellapantula2021deep}. For every dimension $D$ considered, the phase-space contains the first $D$ features mentioned above. The number of data points is $N=8 \times 10^6$. Each physical feature is rescaled in the dataset. 
In the present work, the distribution of the data points is altered in the downsampled dataset. However, this alteration does not affect the progress variable statistics since they are assembled prior to the downsampling process.
An illustration of the two-dimensional version of the dataset is shown in Fig.~\ref{fig:dataset}.

\subsection{Quantitative criterion}

\subsubsection{Definition}
Evaluation of the performance of Algo.~\ref{algo:iterative} was done visually in Sec.~\ref{sec:method}, which is possible in at most two dimensions. To quantitatively evaluate the quality of the downsampled dataset, an intuitive approach is to ensure that the distance of every data point to its closest neighbor is large. This method emphasizes coverage of the phase-space and evaluates whether there exist clusters of data points. In practice, the average distance of each point to its closest neighbor is used, and the distance criterion is expressed as

\begin{equation}
    \text{distance criterion} = \frac{1}{n} \sum_{i=1}^n ||x_i-x_{ni}||_2,
\end{equation}

\noindent where $x_i$ is the $i^{th}$ data point of the reduced dataset and $x_{ni}$ is the nearest neighbor of the $i^{th}$ data point. This metric is referred to as the \textit{distance criterion} in the rest of the manuscript.

\subsubsection{Suitability of the distance metric}
\begin{figure}[h!]
    \centering
    \includegraphics[width=0.30\textwidth]{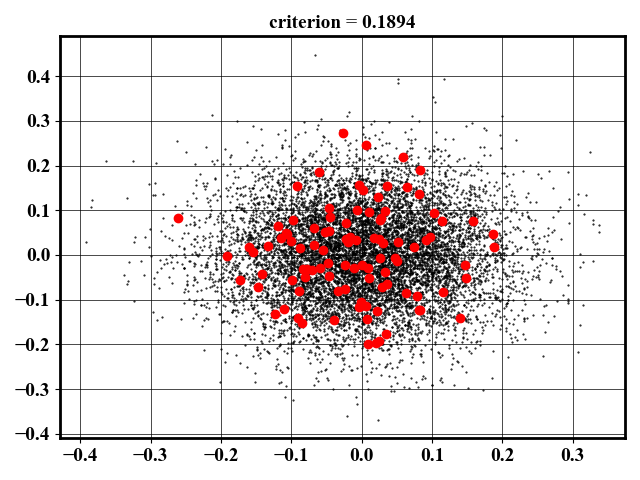}
    \includegraphics[width=0.30\textwidth]{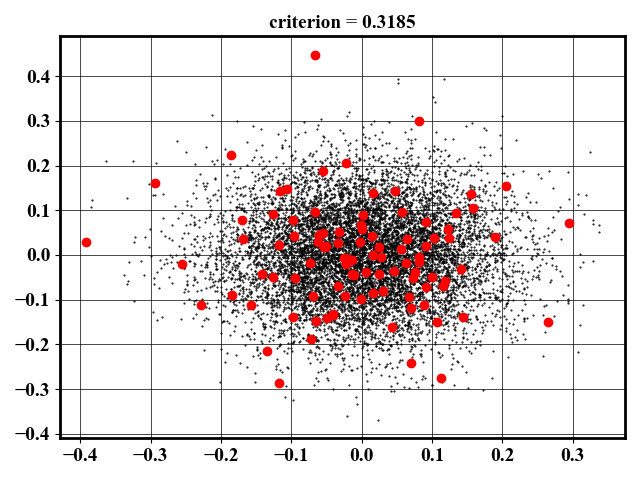}
    \includegraphics[width=0.30\textwidth]{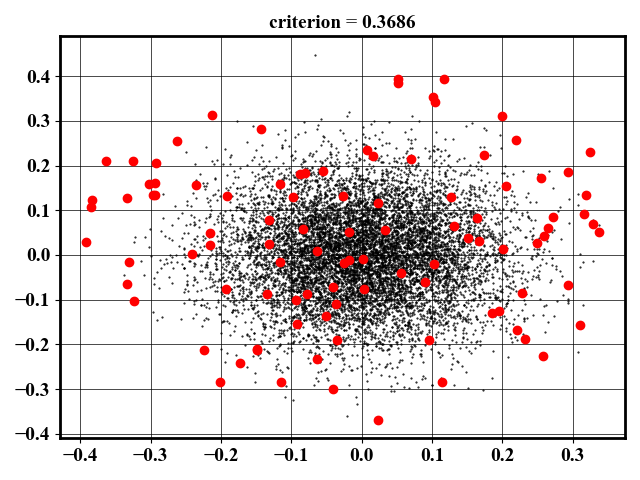}
    \includegraphics[width=0.30\textwidth]{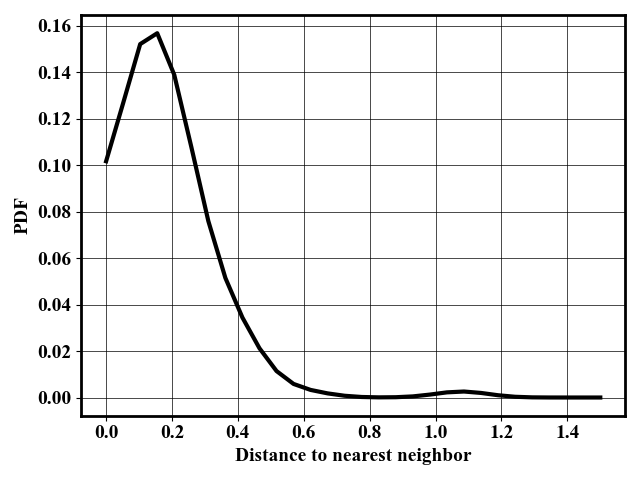}
    \includegraphics[width=0.30\textwidth]{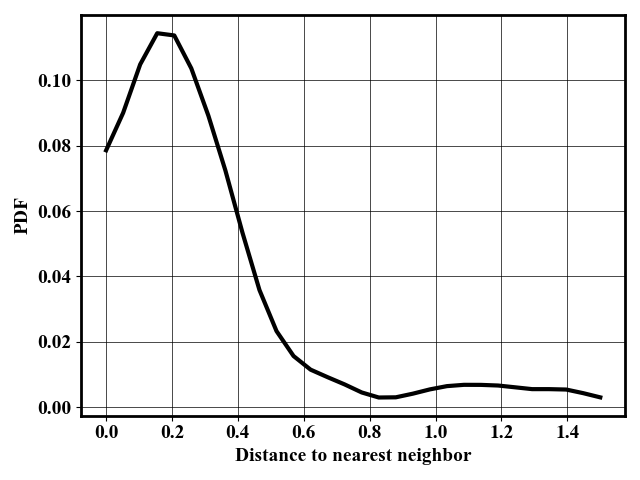}
    \includegraphics[width=0.30\textwidth]{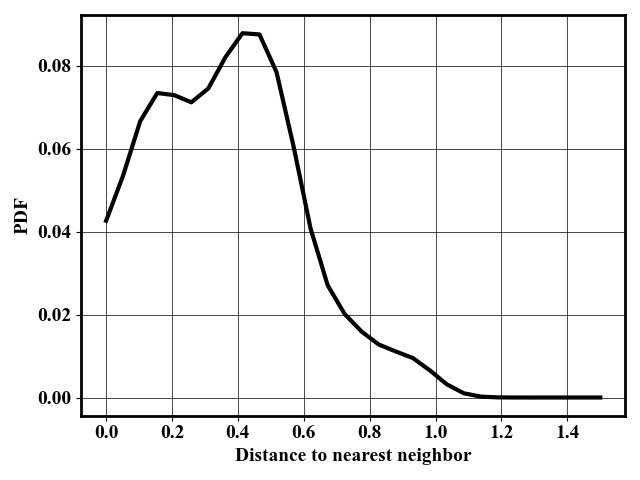}
    \caption{Top: comparison of selection schemes for 100 samples (\mydot{red}) out of $10^5$ data points distributed according to $\mathcal{N}([0,0],diag([0.1,0.1]))$ (\mydot{black}). Bottom: PDF of the distance to nearest neighbor in the downselected dataset. Left: random selection scheme. Middle: brute force optimization of distance criterion. Right: Algo.~\ref{algo:iterative}}
    \label{fig:criterion}
\end{figure}

Starting from an arbitrarily distributed dataset, the objective of the method described in Sec.~\ref{sec:method} is to downselect data points to obtain a uniform distribution in phase-space. In this section, the focus is on assessing whether the samples are distributed as expected. For a two-dimensional phase-space, a visual assessment can be used as in Sec.~\ref{sec:singlepassAlgo} and Sec.~\ref{sec:iterativeAlgo}. In higher dimensions, such an assessment is not possible, and quantitative metrics become necessary. Here, it is proposed to compute the average distance of every data point to its closest neighbor. To make distances along each dimension equivalent, the metric is computed on a rescaled dataset such that each dimension spans the interval $[-4,4]$. The suitability of the metric is assessed with a canonical bivariate normal distribution $\mathcal{N}([0,0], diag([0.1,0.1]))$. The starting dataset comprises $10^4$ samples and is reduced to $10^2$ in the final dataset. Three strategies for selecting the reduced dataset are compared to evaluate whether the criterion chosen coincides with a uniform distribution of data. A random set of samples is shown in Fig.~\ref{fig:criterion} (left) and results in a distance criterion of $0.1894$. The result of a brute force search in the set of $10^2$ data points that maximizes the criterion is shown in Fig.~\ref{fig:criterion} (middle). The brute force search is conducted by repeating a random selection of $10^2$ samples at each iteration ($10^5$ iterations are used) and selecting the set that maximizes the distance criterion. This procedure results in a distance criterion of $0.3185$, and the set of downselected points obtained offers better coverage of phase-space than the random search. Samples obtained from Algo.~\ref{algo:iterative} (Fig.~\ref{fig:criterion}, right) give a distance criterion of $0.3686$. There again, the criterion improved, which qualitatively coincides with a uniform distribution of the downselected points. In this section, it is not claimed that Algo.~\ref{algo:iterative} converges to the dataset that optimizes the distance criterion value. However, it is claimed that improving the distance criterion coincides with uniformly distributing the data points and that the performance of Algo.~\ref{algo:iterative} can be assessed quantitatively using the proposed criterion.

\subsection{Effect of the number of flow iterations}
\label{sec:effectIteration}

The proposed distance criterion is used to illustrate the effect of the number of iterations. The results are shown in Tab.~\ref{tab:criterionVSiteration}. The mean and standard deviation of the distance criterion values obtained with five independent runs using $M=10^5$ are shown. Sampling quality dramatically increases after the first iteration and then stagnates. This result is in line with Fig.~\ref{fig:conditionalError}, where a conditional error on the sampling probability was shown to not improve after two iterations. A consequence of this result is that Algo.~\ref{algo:iterative} increases the computational cost only twofold. Another valuable result is that the standard deviation of the distance criterion decreases when using more than one iteration, which suggests that iterating improves the robustness of the method. 

The results were compared with random IS and stratified sampling \citep{yellapantula2021deep,nguyen2021machine}, where each stratum is a cluster obtained from the k-means algorithm \citep{lloyd1982least}. Different numbers of clusters were tested in the range $[20,160]$. The best results are shown in Tab.~\ref{tab:criterionVSiteration} and were obtained with 40 clusters. Overall, Algo.~\ref{algo:base} and Algo.~\ref{algo:iterative} are superior to random and stratified sampling. Visual inspection of the results (see Appendix~\ref{app:visualInspection}) confirms that using more than two iterations does not help with the sampling quality and that the sampling quality of the proposed method exceeds that of the random and stratified sampling.  Appendix~\ref{app:buildingData} shows the application of Algo.~\ref{algo:iterative} to a different dataset. Two iterations appears to also suffice in this case, which suggests that the aforementioned findings might be applicable to a wide variety of datasets.


\begin{table}[h]
\caption{Dependence of the distance criterion value on the number of iterations and instances selected ($n$). Larger distance criterion value is better. Normalizing flow is used for the PDF estimation.}
\label{tab:criterionVSiteration}
\begin{tabular}{ |c|c|c|c|c|c|c| } 
\hline
 Norm. flow & Algo.~\ref{algo:base} & Algo.~\ref{algo:iterative} (2 iter.) & Algo.~\ref{algo:iterative} (3 iter.)  & Random & Stratified\\
\hline 
 $n=1,000$ & $0.064 \pm 0.014$ & $\boldsymbol{0.098 \pm 0.003}$ & $0.096 \pm 0.003$ & $0.046$ & $0.067$  \\
 \hline
 $n=10,000$ & $0.022 \pm 0.0033$ &  $\boldsymbol{0.030 \pm 0.0002}$ & $0.029 \pm 0.0006$ & $0.014$ & $0.021$ \\
 \hline

\end{tabular}
\end{table}

Table.~\ref{tab:criterionVSiterationBin} shows the results when using a binning strategy for the computation of the PDF, as well as the method of \cite{biswas2018situ} where the PDF is computed with all the available data with a binning strategy. Since the full dataset is used compute the PDF, no iterative procedure is needed. The sampling is done by discretizing each dimension with 100 equidistant bins. The binning strategy with $M=10^5$ does exhibit similar results as the one with normalizing flow. The iterative procedure does improve the computation of the PDF which demonstrate that the predictor-corrector strategy is applicable to other density estimation methods. Finally, the method of \cite{biswas2018situ} does perform as well as the proposed method, albeit requiring to compute the density with the full dataset. Visualization of the effect of iteration with the binning strategy are shown in Appendix~\ref{app:visualInspection}.

\begin{table}[h]
\caption{Dependence of the distance criterion value on the number of iterations and instances selected ($n$). Larger distance criterion value is better. Binning is used for the PDF estimation.}
\label{tab:criterionVSiterationBin}

\begin{tabular}{ |c|c|c|c|c| } 
\hline
 Binning & Algo.~\ref{algo:base} & Algo.~\ref{algo:iterative} (2 iter.) & Algo.~\ref{algo:iterative} (3 iter.)&  \cite{biswas2018situ}  \\
 \hline 
  $n=1,000$ & $0.069 \pm 0.0005$ & $0.091 \pm 0.0099$ & $\boldsymbol{0.099 \pm 0.0023}$ &   $0.097 \pm 0.0011$  \\
 \hline
  $n=10,000$ & $0.019 \pm 0.0001$ & $\boldsymbol{0.030\pm 0.0002}$ &  $0.030\pm 0.0003$ & $\boldsymbol{0.031 \pm 0.0002}$ \\
 \hline

\end{tabular}

\end{table}

\begin{figure}[h!]
    \centering
    \includegraphics[width=0.47\textwidth]{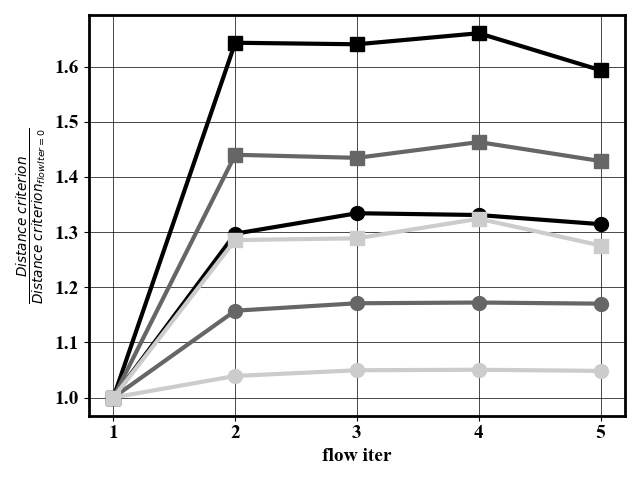}
    \caption{Sensitivity of distance criterion to the number of flow iterations, normalized by the distance criterion after the first flow iteration. Symbols represent the mean distance criterion. Results are shown for $D=2$ (\mythickbarredcircle{black}{black}) and $D=4$ (\mythickbarredsquare{black}{black}), for $n=10^3$ (\mythickline{black}), $n=10^4$ (\mythickline{black!60}), and $n=10^5$ (\mythickline{black!20})}
    \label{fig:effectIterationHighDim}
\end{figure}

In high dimensions, the same conclusions can be obtained. Figure~\ref{fig:effectIterationHighDim} shows the average distance criterion obtained over five independent runs, for $D$ varying in the set $\{2,4\}$ and $n$ varying in the set $\{ 10^3, 10^4, 10^5 \}$. The distance criterion is normalized by the value reached at the first iteration to quantify the benefit of iterating. For $D=2$ and $D=4$, the quality of the downsampled dataset dramatically improves after two iterations and then ceases to improve. It also appears that iterating mostly benefits aggressive data reduction (small $n$) and cases with more dimensions.

\subsection{Sampling quality in high dimension}
\label{sec:quality}

Here, the distance criterion is used to assess the sampling quality in higher dimensions. The distance criterion obtained with Algo.~\ref{algo:iterative} is shown for different phase-space dimensions using $M=10^5$. Random sampling and stratified sampling are used to compare the obtained sampling quality. The stratified sampling procedure was performed with the optimal number of clusters found in Sec.~\ref{sec:effectIteration}. The sampling results are shown in Tab.~\ref{tab:highDimCriterion}.

\begin{table}[h]
\caption{Distance criterion values for different numbers of instances selected ($n$) and data dimensions ($D$). Larger distance criterion value is better. Normalizing flow is used for the PDF estimation. Data in parenthesis are results with $M = 4\times 10^5$}.
\label{tab:highDimCriterion}
\begin{tabular}{ |c|c|c|c|c| } 
\hline
 Norm. Flow & Algo.~\ref{algo:base} & Algo.~\ref{algo:iterative} ($M=10^5$, 2 iter.) & Random & Stratified \\
 \hline 
  $n=1,000$ $D=3$ & $0.094 \pm 0.0161$  & $0.109 \pm 0.0132$   & $0.065 \pm 0.0012$ &  $0.095 \pm 0.0030$  \\
  &   & ($\boldsymbol{0.136 \pm 0.0032}$) & & \\
 \hline
  $n=1,000$ $D=4$ & $0.159 \pm 0.0267$ & $\boldsymbol{0.223 \pm 0.0129}$ & $0.106 \pm 0.0048$ & $0.137 \pm 0.0034$    \\
 \hline
 $n=1,000$ $D=5$ & $0.282 \pm 0.0244$ & $\boldsymbol{0.330 \pm 0.0036}$ & $0.133 \pm 0.0031$ & $0.188 \pm 0.0064$    \\
 \hline 
  $n=10,000$ $D=3$ & $0.034 \pm 0.0083$ & $0.041\pm 0.0044$ & $0.026 \pm 0.0003$ & $0.038 \pm 0.0005$    \\
  &  & ($\boldsymbol{0.057 \pm 0.0006}$) & & \\
 \hline
  $n=10,000$ $D=4$ & $0.084 \pm 0.0028$ & $\boldsymbol{0.098\pm 0.0064}$ & $0.049 \pm 0.0004$ & $0.065 \pm 0.0014$    \\
 \hline
 $n=10,000$ $D=5$ & $0.166 \pm 0.0073$ & $\boldsymbol{0.173\pm 0.0061}$ & $0.068 \pm 0.0009$ & $0.094 \pm 0.0059$    \\
 \hline
\end{tabular}
\end{table}

\begin{figure}[h!]
    \centering
    \includegraphics[width=0.30\textwidth]{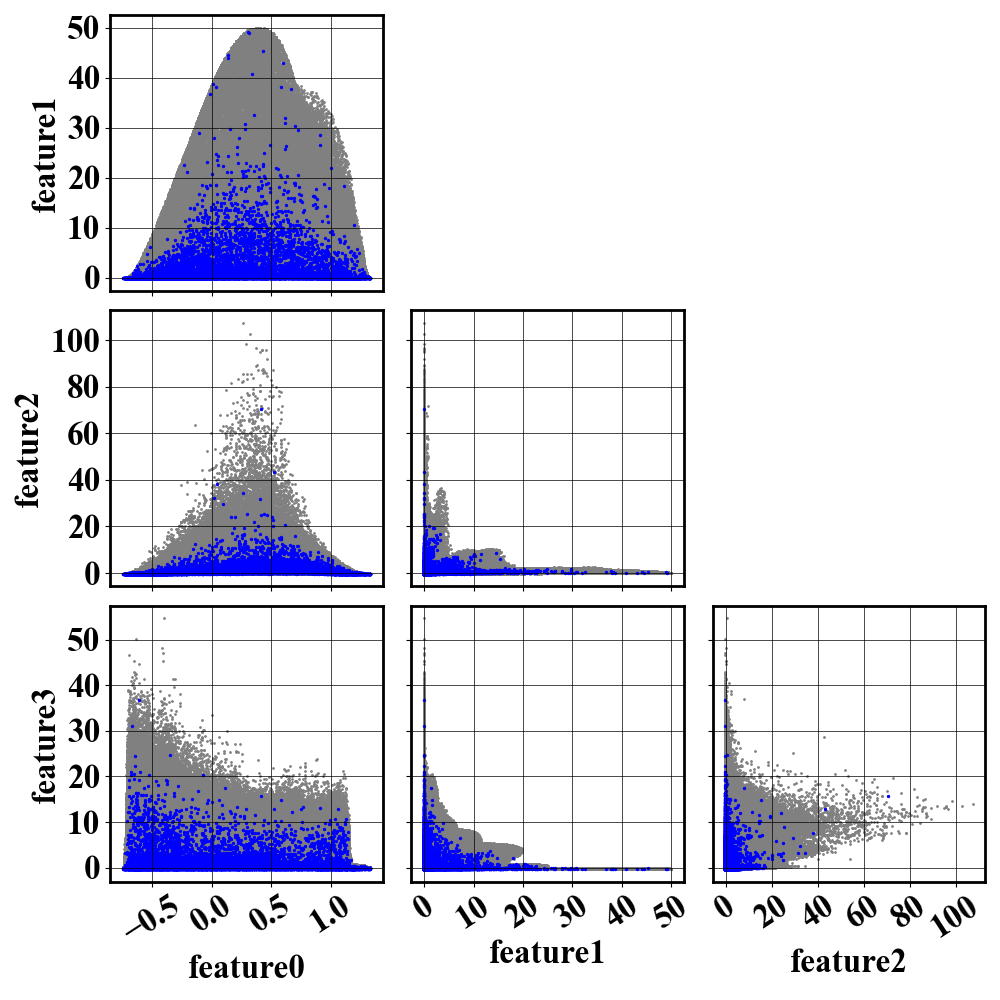}
    \includegraphics[width=0.30\textwidth]{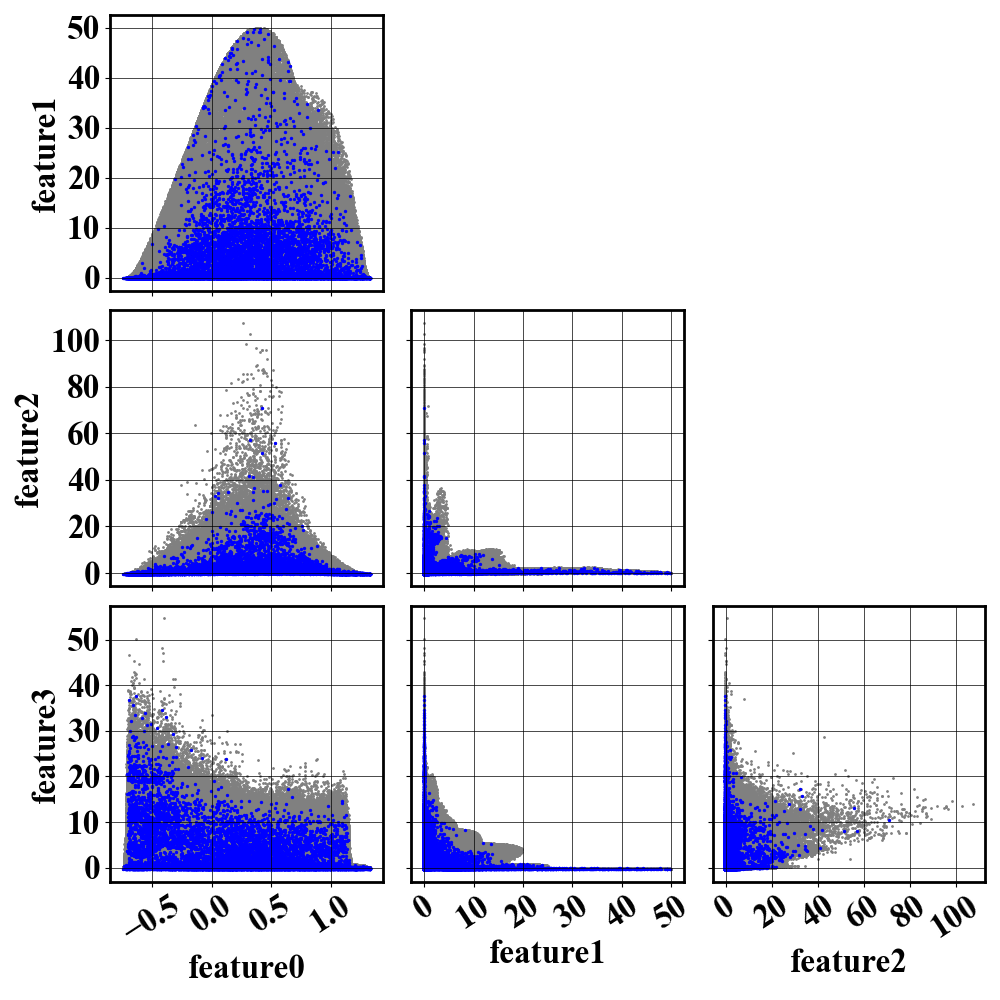}
    \includegraphics[width=0.30\textwidth]{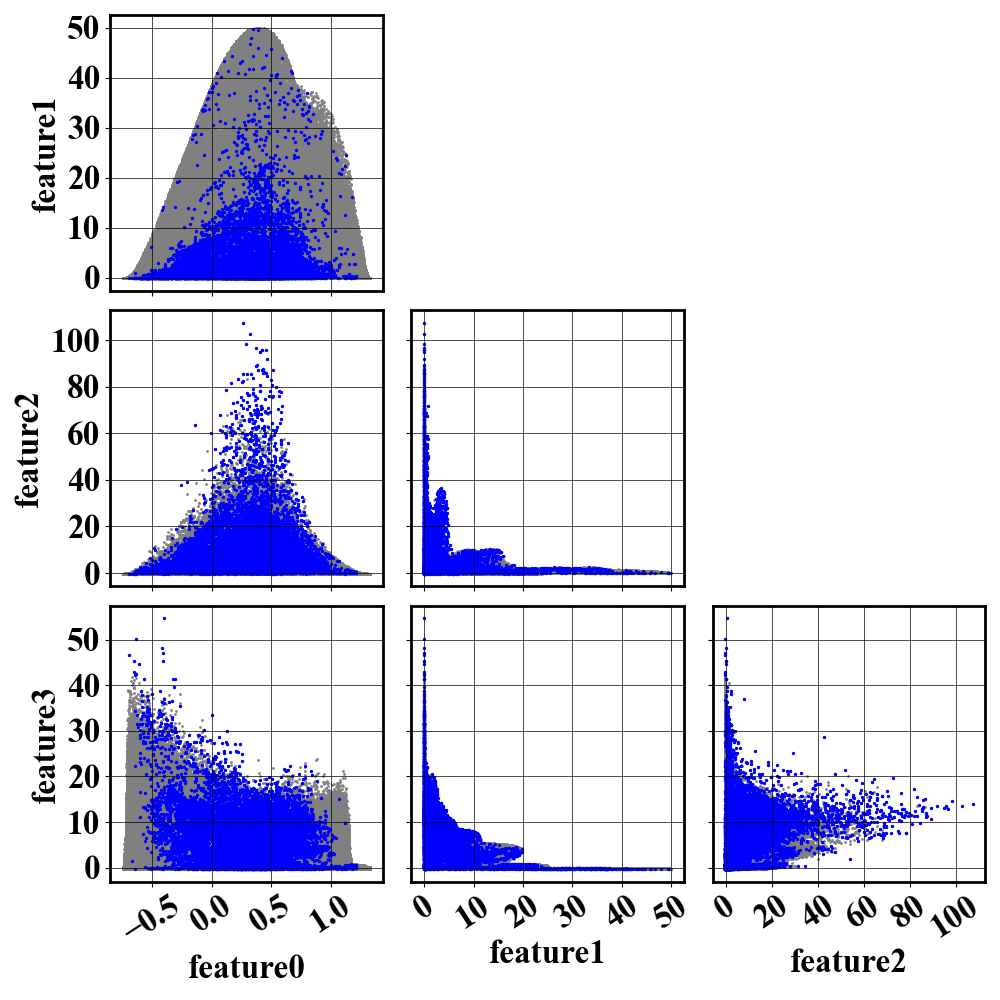}
    \caption{Two-dimensional projection of $n=10,000$ selected data points (\mydot{blue}) out of an original four-dimensional dataset (\mydot{gray}). Left: random sampling. Middle: stratified sampling. Right: Algo.~\ref{algo:iterative}}
    \label{fig:highDimCornerPlot}
\end{figure}

As can be seen in Tab.~\ref{tab:highDimCriterion}, the proposed sampling technique also performs better than traditional methods in more than two dimensions. To further verify this conclusion, a two-dimensional projection of the data along every dimension pair is shown for the three methods for $D=4$ and $n=10,000$ in Fig.~\ref{fig:highDimCornerPlot}. Visually, the proposed method better covers the phase-space. The data points selected by Algo.~\ref{algo:iterative} cover the phase-space differently than when only the first two features are considered (Fig.~\ref{fig:highDimCornerPlot}, right). This observation can be explained by the other two-dimensional projections; variations in $\widetilde{\chi}_{res}$, $\alpha$, and $\beta$ in the dataset are mostly observed at low $\widetilde{C^{''2}}$ values.

\begin{table}[h]
\caption{Distance criterion values for different numbers of instances selected ($n$) and data dimensions ($D$). Larger distance criterion value is better. Binning is used for the PDF estimation.}
\label{tab:highDimCriterionBin}

\begin{tabular}{ |c|c|c|c| } 
\hline
 Binning & Algo.~\ref{algo:base} & Algo.~\ref{algo:iterative} ($M=10^5$, 2 iter.) & \cite{biswas2018situ}  \\
 \hline 
  $n=1,000$ $D=3$ & $0.124 \pm 0.0038$ & $0.133 \pm 0.0021$ & $\boldsymbol{0.151 \pm 0.0015}$    \\
 \hline
  $n=1,000$ $D=4$ & $0.183 \pm 0.0023$ & $0.206 \pm 0.0004$  & $\boldsymbol{0.233 \pm 0.0019}$ \\
 \hline
 $n=1,000$ $D=5$ & Out Of Memory & Out Of Memory & Out Of Memory   \\
  \hline 
  $n=10,000$ $D=3$ & $0.047 \pm 0.0001$ & $0.046 \pm 0.001$ & $\boldsymbol{0.063 \pm 0.0003}$    \\
 \hline
  $n=10,000$ $D=4$ & $0.083 \pm 0.0005$ & $0.096 \pm 0.0006$  & $\boldsymbol{0.112 \pm 0.0004}$ \\
 \hline
 $n=10,000$ $D=5$ & Out Of Memory & Out Of Memory & Out Of Memory   \\
  \hline 
\end{tabular}
\end{table}

Similar to the two dimension case, Tab.~\ref{tab:highDimCriterionBin} shows that the binning strategy also benefits from the predictor-corrector algorithm, which further suggests that Algo.~\ref{algo:iterative} can benefit density estimation methods in general. For the case $D=3$ the binning strategy with $M=10^5$ outperforms normalizing flows. After further investigation, it appeared that using $M=4 \times 10^5$ significantly improved the case $D=3$ with normalizing flow (the distance criterion values are shown in parenthesis in Tab.~\ref{tab:highDimCriterion}). The sensitivity analysis with $M$ shown in Fig.~\ref{fig:sensitivityM} also suggests that the case $D=3$ gives poor results with $M<4 \times 10^5$. Note that for other higher dimensions, similar considerations also apply but have a lesser impact than for $D=3$. The method of \cite{biswas2018situ} outperforms the present method for $D<5$, at the expense of higher memory requirements and computing the PDF with the full dataset. For $D \geq 5$, all binning strategies fail due to too large memory requirements.

\subsection{Data size limits}
\label{sec:sensitivityM}

\begin{figure}[h!]
    \centering
    \includegraphics[width=0.47\textwidth]{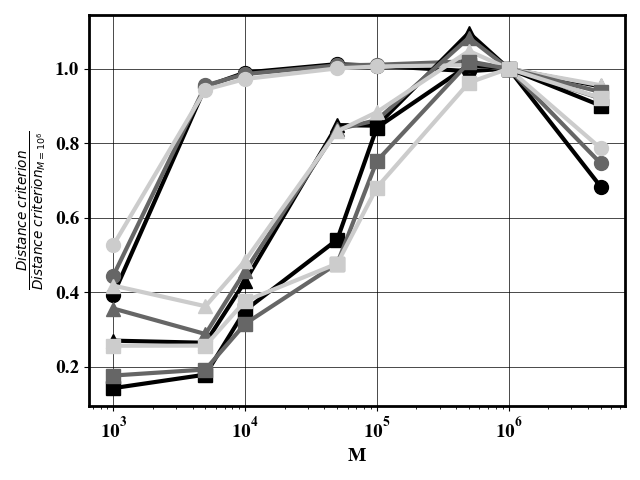}
    \caption{Sensitivity of distance criterion to $M$ normalized by the distance criterion at $M=10^6$. Symbols represent the mean distance criterion. Results are shown for $D=2$ (\mythickbarredcircle{black}{black}), $D=3$ (\mythickbarredtriangle{black}{black}), and $D=4$ (\mythickbarredsquare{black}{black}), for $n=10^3$ (\mythickline{black}), $n=10^4$ (\mythickline{black!60}), and $n=10^5$ (\mythickline{black!20})}
    \label{fig:sensitivityM}
\end{figure}

To construct the probability map efficiently, it is proposed in Sec.~\ref{sec:method} training a normalizing flow on a subset of the full dataset. In addition, to account for rare realizations, it was proposed to estimate the probability map iteratively (Sec.~\ref{sec:iterativeAlgo}). In this section, it is investigated how large the data subset needs to be to still obtain a uniform-in-phase-space reduced -dataset. In other words, the sensitivity of Algo.~\ref{algo:iterative} to $M$ is evaluated. The dataset presented in Sec.~\ref{sec:singlepassAlgo} is considered. Algorithm~\ref{algo:iterative} is used with two iterations with a feature space of dimension $D$ in the set $\{2, 3, 4\}$. The original dataset ($N=8 \times 10^6$) is downsampled to $n$ data points in the set $\{ 10^3, 10^4, 10^5 \}$. For each case, $M$ is varied in the set $\{ 10^3, 5 \times 10^3, 10^4, 5 \times 10^4, 10^5, 5 \times 10^5, 10^6, 5 \times 10^6 \}$. Each data reduction is performed five times, and the ensemble averages of the distance criterion are shown in Fig.~\ref{fig:sensitivityM}. The number of training steps is fixed at $12,000$ (the number of epochs varies between $6$ and $12,000$) to ensure that larger $M$ values do not result in worse convergence of the normalizing flow.

Figure~\ref{fig:sensitivityM} shows that Algo.~\ref{algo:iterative} does not favor small or high $n$ values and that varying $M$ equally affects cases with small or large $n$. The value of $M$ only affects the accuracy of the probability map, which has equal implications for all sizes of downsampled data. 

In the limit case where $M$ is small, the distance criterion is, across the board, lower than the optimal value. This can be understood by considering an asymptotic scenario where $M=1$. In that case, the best possible probability map will be a delta distribution centered on the data point used. At the next flow iteration, the probability map will be corrected with another delta distribution. A very small value of $M$ inherently limits the expressiveness of the normalizing flow, i.e., the complexity of the probability map that can be approximated, thereby adversely affecting the reduced dataset. $M$ needs to be sufficiently high to formulate a first reasonable approximation of the probability map that can be accurately corrected at the next flow iteration. 

As $M$ increases, the distance criterion ceases to improve, which confirms that a subset of the full dataset can be used to train the normalizing flow. The higher the phase-space dimension $D$, the higher the smallest $M$ value required to achieve uniform-in-phase-space sampling. As the dimension increases, the probability map becomes more difficult to approximate, requiring a larger dataset size.

For very large $M$ (close to the size of the full dataset $N$), the performance of the algorithm degrades across the board. This effect can be explained by again considering the asymptotic limit ($M=N$). In the ideal case, the probability map is already approximated at the first iteration. At the second iteration, a correction is applied if the true probability map is not uniform, which is the case for most datasets.

If the optimal $M$ value depends on the complexity of the probability map to be approximated, it also depends on the particular dataset considered. Because the performances of the method vary smoothly with $M$, it would be possible, in practice, to try multiple $M$ values until an optimum is reached.

\subsection{Computational cost}
\label{sec:computationalcost}

Compared to random or stratified sampling, Algo.~\ref{algo:iterative} is more expensive, as a probability needs to be estimated and evaluated for every data point. The computational cost of the method and its scale-up with the number of data points and dimensions is described here.

\begin{figure}[h!]
    \centering
    \includegraphics[width=0.30\textwidth]{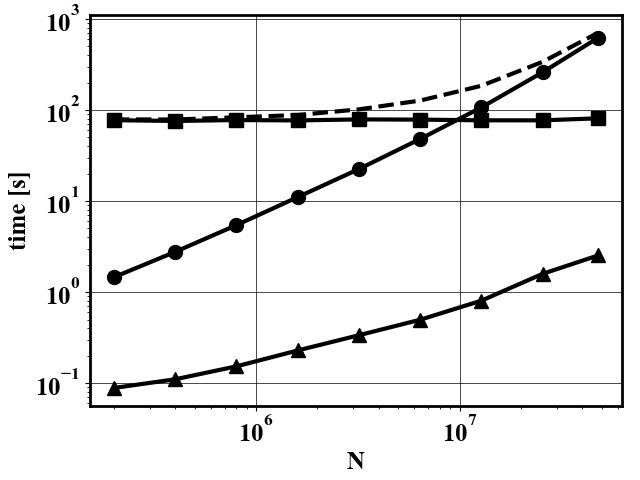}
    \includegraphics[width=0.30\textwidth]{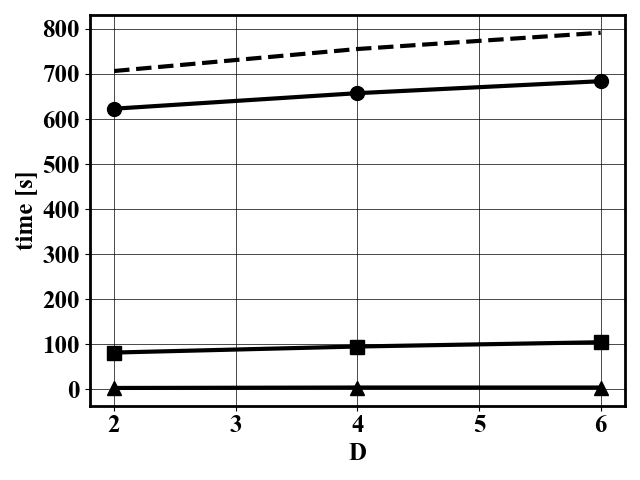}
    \caption{Scaling of computational cost against the number of data points $N$ in the full dataset (left) and the dimension $D$ (right). The computational cost is divided into the probability map estimation (\mybarredsquare{black}{black}), called Step 1; the construction of acceptance probability (\mybarredcircle{black}{black}), called Step 2a; and the data selection (\mybarredtriangle{black}{black}), called Step 2b. The total computational cost is also indicated for both scaling plots (\mythickdashedline{black})}
    \label{fig:scalingNd}
\end{figure}

The computational cost of the proposed method can be divided into three separate components. These components (referred to as Steps) are the building blocks of Algo.~\ref{algo:base}, which is itself the building block of Algo.~\ref{algo:iterative}. First, the predictive model for the probability of the data points as a function of the location in phase-space is trained (Step 1). Second, the predictive model is evaluated for each data point in the dataset (Step 2a). Third, the acceptance probability for each data point is constructed, and the data points of the reduced dataset are selected (Step 2b). The computational cost study is done with the dataset $D_1$ \citep{savard2017effects,lapointe2015differential} used in \citet{yellapantula2021deep}, which is different than the dataset considered in Sec.~\ref{sec:dataset} in that it contains more data points ($N=4.8\times 10^7$). The larger number of data points allows to clearly distinguish regimes where the cost of Step 1 or Step 2a dominates. The scaling with respect to $N$ is obtained with $D=2$. The scaling with respect to $D$ is obtained with $N=4.8\times 10^7$. The breakdown of the computational cost of Algo.~\ref{algo:iterative} is shown in Fig.~\ref{fig:scalingNd} as a function of $N$ (left) and $D$ (right). 

The computational cost of Step 1 mainly depends on the number of data points used to train the model. As seen in Sec.~\ref{sec:method}, a subsample of the full dataset can be used to train the model. No matter how many data points are in the original dataset, the predictor-corrector algorithm (Algo.~\ref{algo:iterative}) allows the model to be efficiently trained with a small number of data points. Therefore, as can be observed in Fig.~\ref{fig:scalingNd} (left), the computational cost of Step 1 does not depend on the total number of data points $N$. The complexity of Step 1 is difficult to rigorously define as it depends of the convergence of the optimization. Typically, it will scale linearly with $M$ if the number of epochs is kept constant for different $M$ values. The number of dimensions $D$ influences the number of parameters in the network as the input space and the output space depend on $D$. In turn, the training time linearly varies with the number of trainable parameters. In the present case the number of parameters increased as $D^{1/4}$. In our case, the time complexity of Step 1 can be estimated as $\mathcal{O}(MD^{1/4})$.

The computational cost of Step 2a linearly scales with $N$ as the probability model needs to be evaluated for every data point. The computational cost also linearly scales with the number of trainable parameters used in the normalizing flow as each probability evaluation requires a neural-network inference. The complexity of Step 2a can be estimated as $\mathcal{O}(ND^{1/4})$.

Finally, Step 2b scales as $\mathcal{O}(N)$, but its contribution to the total computational cost is negligible compared with Step 2a.

\begin{figure}[h!]
    \centering

    \includegraphics[width=0.30\textwidth]{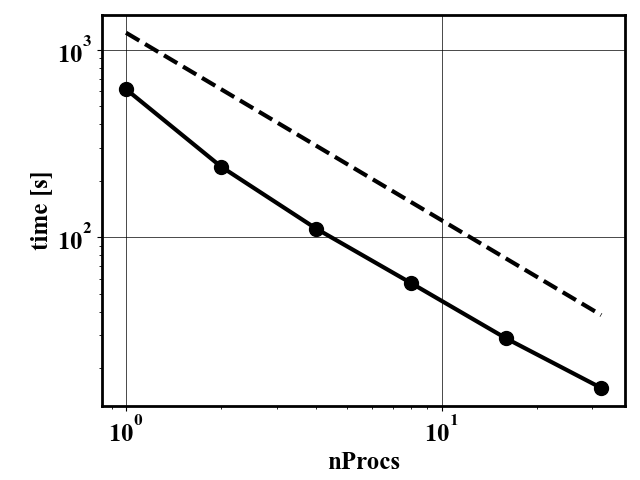}
    \includegraphics[width=0.30\textwidth]{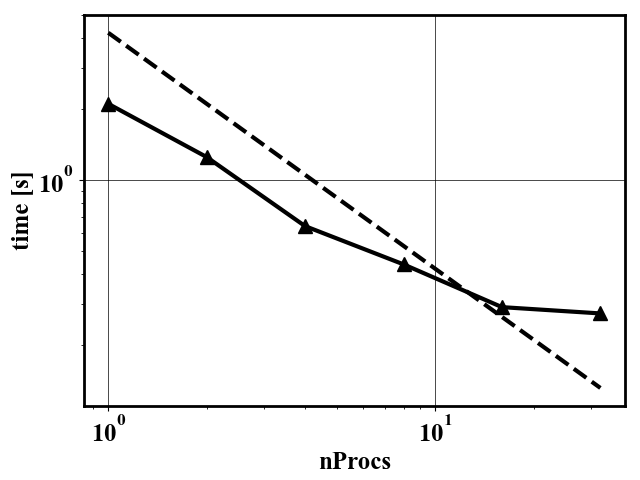}
    \caption{Left: scaling of Step 2a (\mybarredcircle{black}{black}) overlaid with linear scaling (\mythickdashedline{black}). Right: weak scaling of Step 2b (\mybarredcircle{black}{black}) overlaid with linear scaling (\mythickdashedline{black})}
    \label{fig:scalingNProc}
\end{figure}

In Fig.~\ref{fig:scalingNd}, it can be observed that for large datasets, the total cost mainly depends on Step 2a. However, since Step 2a solely consists of probability density evaluations, it is embarrassingly parallelizable. With a simple MPI-parallelization, the computational time of Step 2a can be considerably decreased. The weak scaling plot of Step 2a is shown in Fig.~\ref{fig:scalingNProc}. Here, memory requirements are driven by the data points stored in memory, rather than by the probability density model (as would be the case if one stored it as a high-dimensional histogram). Therefore, the MPI-parallelization also decreases the memory requirements, which is beneficial for very large datasets. In addition, although Step 2b (adjustment of the acceptance probability to reach the desired data reduction) is not embarrassingly parallel, the global summation of acceptance probability can easily be parallelized by partitioning the data (see Fig.~\ref{fig:scalingNProc}, right). The parallelized implementation of Step 2a and Step 2b is also made available in the companion repository. 

In conclusion, the cost of the procedure linearly scales with the number of data points and can be significantly driven down if executed on a parallel architecture. Step 2a and 2b can be accelerated via MPI-parallelization. Step 1 can be accelerated by a judicious choice of the number of data points used to train the normalizing flow. In addition, because Step 1 consists of training a neural net, it can benefit from GPU-acceleration for large batches or networks (which is also implemented in the companion repository).

\section{Application: data-efficient ML}
\label{sec:dataEfficient}

In this section, the effect of the uniform selection of data in phase-space is assessed for the training of data-driven models. This data selection method is compared to random sampling and stratified sampling on a realistic scientific dataset. The input-output mapping in this dataset is noisy because the input dimensions are insufficient to perfectly model the output. Thus, a synthetic dataset where the noise level can be controlled is considered to understand the resilience of the method to noise. 

\subsection{Example of closure modeling}

\begin{figure}[h!]
    \centering
    \includegraphics[width=0.4\textwidth]{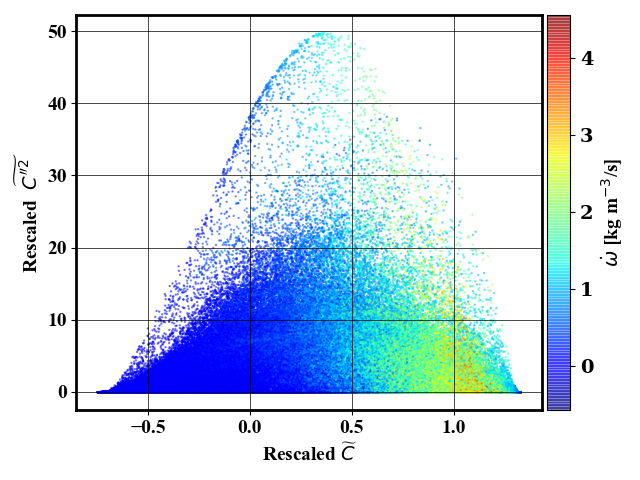}
    \caption{Illustration of the combustion dataset. The dots are colored by progress variable source term $\dot{\omega}$}
    \label{fig:srcProgDataset}
\end{figure}

Typically, combustion problems require solving partial differential equations for the transport of species mass fractions or some representative manifold variables, such as the progress variable \citep{pierce2004progress}. In large eddy simulations of turbulent combustion problems, a closure model is required for the filtered reaction rate of such variables. One strategy consists of using data available via DNS to formulate a data-driven model for the unclosed terms. For example, in premixed flames, it may be useful to transport a filtered progress variable $\widetilde{C}$ and its variance $\widetilde{C^{''2}}$, and to model the source term of the progress variable transport equation $\dot{\omega}$ as a function of $\widetilde{C}$ and $\widetilde{C^{''2}}$. The functional form of the model would be obtained from an existing dataset. In this section, the objective is to learn such a model. The model is trained on the reduced versions of the dataset used in Sec.~\ref{sec:quality}. The full dataset is illustrated in Fig.~\ref{fig:srcProgDataset}. From this figure, it is apparent that the output quantity is a noisy function of the inputs, indicating that any model formulated will be affected by aleatoric errors.

The IS algorithm proposed in Sec.~\ref{sec:method} can, in principle, be applied for any data-driven model. Two approaches are used here: a Gaussian process \citep{scikit-learn} and a feed-forward neural network \citep{abadi2016tensorflow} with two hidden layers of $32$ units each. The learning rate is initially set at $10^{-1}$ and exponentially decays to $10^{-3}$. The batch size is fixed at $100$, and the number of epochs is $400$. The Gaussian process kernel consists of white noise superimposed with a radial basis function (RBF), where the length scale of the RBF and the standard deviation of the noise are optimized at every realization. The length scale of the RBF is optimized in the interval [$10^{-3}$,$1$] and the noise level is optimized in the interval [$10^{-9}$,$10^5$]. The results shown can be reproduced with the code available in the companion repository. The original dataset contains about $N=8 \times 10^6$ data points. The model training is done on reduced datasets that contain $n=1,000$ or $n=10,000$ data points. The data reduction is done with either Algo.~\ref{algo:iterative} ($M=10^5$ and two flow iterations), random sampling, or stratified sampling. For computational tractability of the Gaussian process, only cases with $n=1,000$ are shown. For every case, the sampling and training are done five times to collect the metric statistics. Two metrics are used here: the mean absolute error and the maximal error over the full dataset ($N$ data points).

\begin{figure}[h!]
    \centering
    \includegraphics[width=0.47\textwidth]{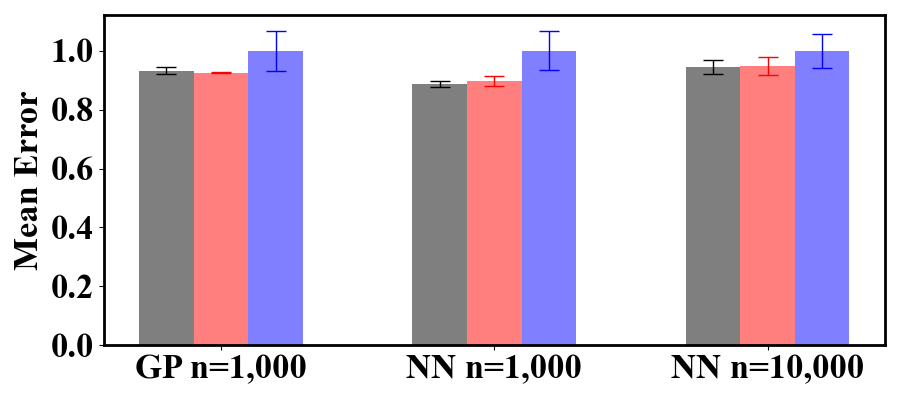}
    \includegraphics[width=0.47\textwidth]{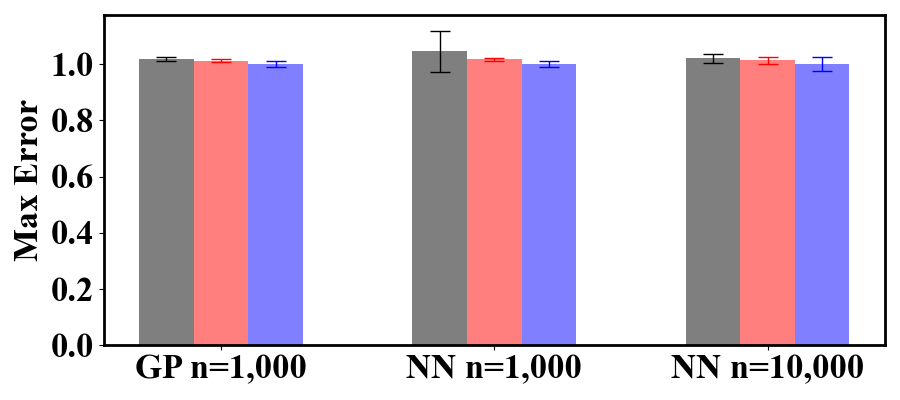}
    \caption{Statistics of the mean (left) and maximal error (right) obtained over five repetitions of downselection and training of a neural network and gaussian process on the combustion dataset for random sampling (\mythickline{black}), stratified sampling (\mythickline{red}), and Algo.~\ref{algo:iterative} (\mythickline{blue}). Bar height shows ensemble mean and error bar shows standard deviation. Bar and error bar size are rescaled by the ensemble mean obtained with Algo.~\ref{algo:iterative}}
    \label{fig:GPNNResultsSRCPROG}
\end{figure}

The training results for the Gaussian process and the feed-forward neural network are shown in Fig.~\ref{fig:GPNNResultsSRCPROG}. Despite covering the phase-space better than the stratified and random sampling, the maximal error obtained by using Algo.~\ref{algo:iterative} is only slightly better than other methods (about 1\% improvement compared to stratified sampling and 2\% compared to random sampling). Furthermore, the average absolute error obtained via Algo.~\ref{algo:iterative} can be slightly larger (by about 6\%) than when using plain random sampling. 

\begin{figure}[h!]
    \centering
    \includegraphics[width=0.31\textwidth]{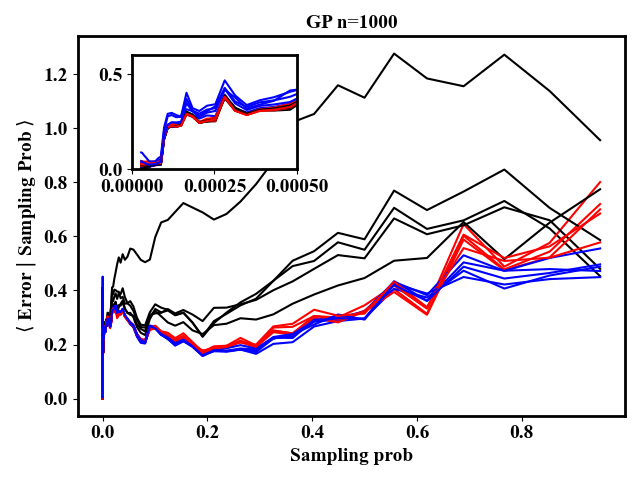}
    \includegraphics[width=0.31\textwidth]{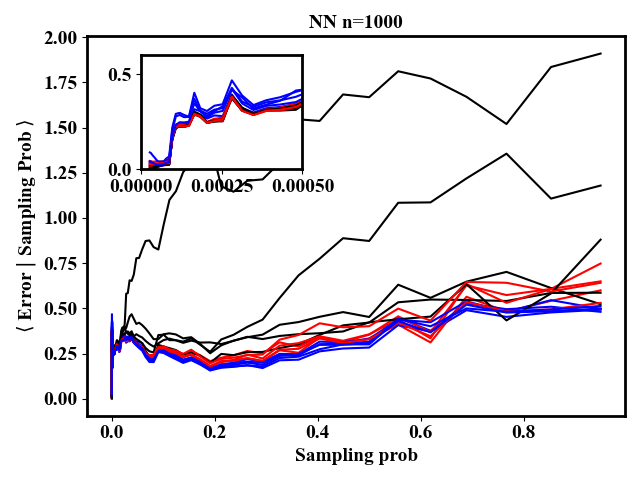}
    \includegraphics[width=0.31\textwidth]{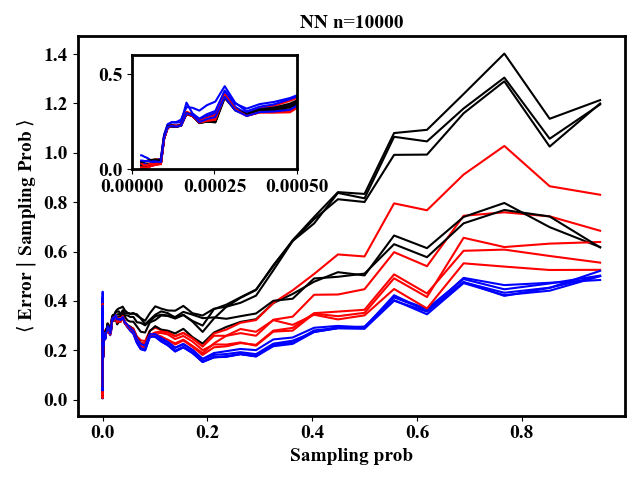}
    \caption{Prediction errors over the full dataset conditioned on sampling probability, calculated with $n=10^4$ for five independent datasets constructed with random sampling (\mythickline{black}), stratified sampling (\mythickline{red}), and Algo.~\ref{algo:iterative} (\mythickline{blue}). Inset zooms in on low sampling probability. Left: Gaussian process with $n=10^3$. Middle: Feed-forward neural network with $n=10^3$. Right: Feed-forward neural network with $n=10^4$}
    \label{fig:conditionalErrorComb}
\end{figure}

To further investigate how the effect of phase-space coverage on the training results, the conditional error obtained with the different sampling procedures is plotted against the sampling probability computed for the $n=10,000$ for five training realizations with all three sampling and training techniques (Fig.~\ref{fig:conditionalErrorComb}). As a reminder, sampling probability is inversely proportional to data density; thus, low sampling probabilities denote high data density regions, and high sampling probabilities denote rare events. For all cases, Algo.~\ref{algo:iterative} induces the lowest errors in rare regions of phase-space (high sampling probability). Stratified sampling, which increases the occurrence of rare events compared with random sampling, also reduces errors compared to random sampling, albeit to a smaller extent than Algo.~\ref{algo:iterative}. However, in high-probability regions of phase-space (small sampling probability), Algo.~\ref{algo:iterative} induces the largest errors (see inset), especially for low values of $n$. Interestingly, Algo.~\ref{algo:iterative} also leads to the least amount of variation when repeating training runs, which indicates robustness to randomness in the construction of the dataset. For metrics such as the mean-squared error, errors in high-probability regions of phase-space dominate, which explains the poor performance of Algo.~\ref{algo:iterative} observed in Fig.~\ref{fig:GPNNResultsSRCPROG}.

Another explanation for the marginal improvement shown in Fig.~\ref{fig:GPNNResultsSRCPROG} is the presence of noise. The core assumption of Algo.~\ref{algo:iterative} is that data points close in phase-space are redundant. However, in presence of noise, redundancy is necessary to correctly approximate the optimal estimator that is being sought. The lower performance of Algo.~\ref{algo:iterative} in high-probability regions compared to random and stratified sampling indicates that the high-probability region of phase-space is also noisy, and thereby benefits from methods that oversample data. The effect of noise is investigated in the next section.

\subsection{Synthetic dataset}

\begin{figure}[h!]
    \centering
    \includegraphics[width=0.9\textwidth]{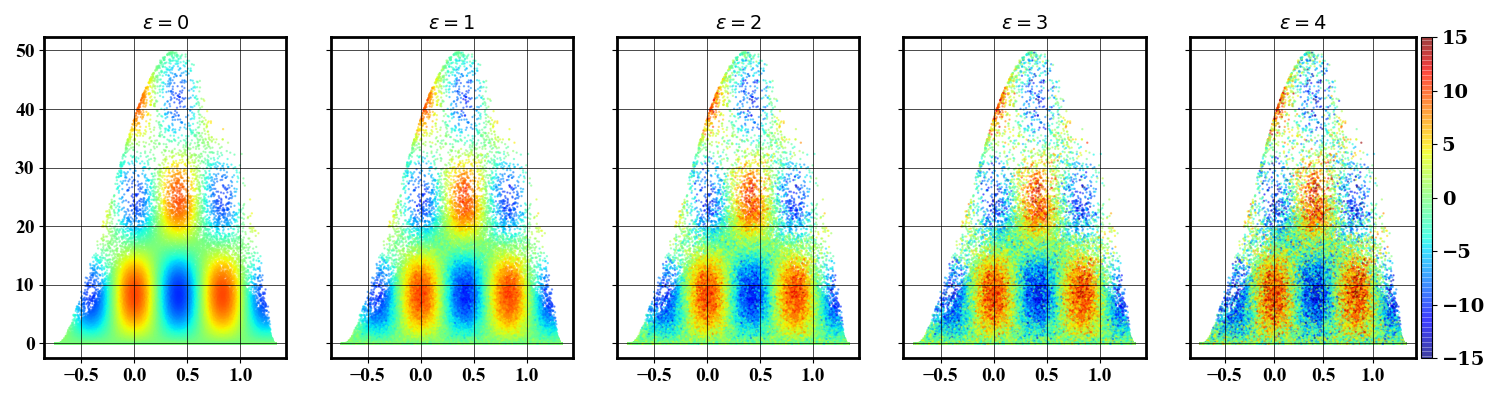}
    \caption{Illustration of the synthetic datasets. Noise level increases from left to right}
    \label{fig:syntheticDataset}
\end{figure}

To better understand the effect of noise on the performance of Algo.~\ref{algo:iterative}, a synthetic dataset is considered. The variables of the phase-space are kept the same, but the variable to learn is replaced by two-dimensional sinusoids shown in Fig.~\ref{fig:syntheticDataset}. The advantage of the synthetic dataset is that one can control the amount of noise to evaluate its effect on the training results. Here, the two-dimensional sinusoids are superimposed with a randomly distributed noise of standard deviation $\varepsilon$ in the set $\{0,1,2,3,4\}$. The functional form of the data is
\begin{equation}
    10 \cos(f_1 \omega_{f_1}) \sin(f_2 \omega_{f_2}) + \eta,
\end{equation}

\noindent where $f_1$ denotes the first feature direction, $f_2$ denotes the second feature direction, $\omega_{f_1}$ is chosen such that $2.5$ oscillation periods span the first feature direction, and $\omega_{f_2}$ is chosen such that $1.5$ oscillation periods span the second feature direction, and $\eta$ is the additive noise.

\begin{figure}[h!]
    \centering
    \includegraphics[width=0.47\textwidth]{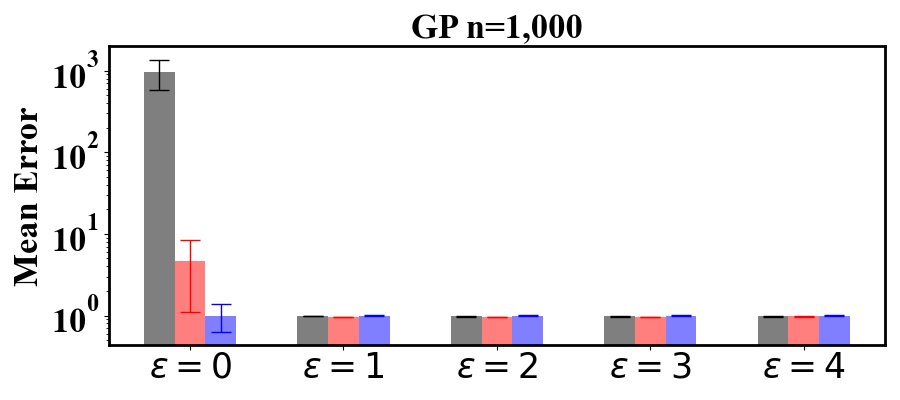}
    \includegraphics[width=0.47\textwidth]{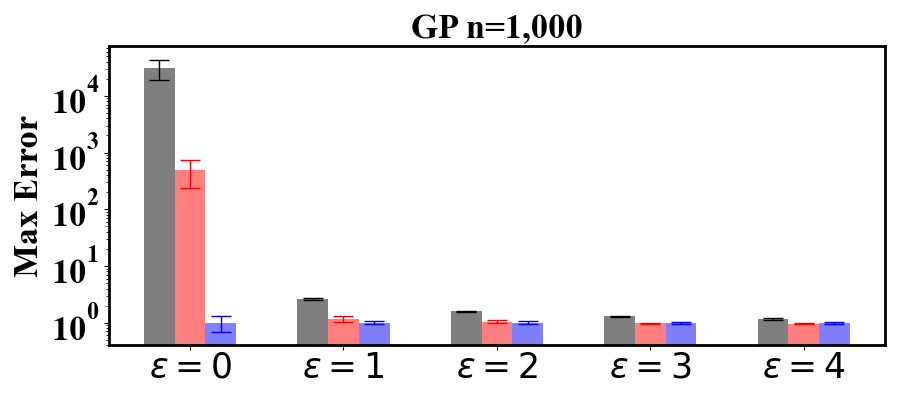}
    \caption{Statistics of the mean (left) and maximal error (right) obtained over five repetitions of downselection and training of a Gaussian process on the synthetic dataset for random sampling (\mythickline{black}), stratified sampling (\mythickline{red}), and Algo.~\ref{algo:iterative} (\mythickline{blue}). Bar height shows ensemble mean and error bar shows standard deviation. Bar and error bar size are rescaled by the ensemble mean obtained with Algo.~\ref{algo:iterative}}
    \label{fig:GPResults}
\end{figure}

\begin{figure}[h!]
    \centering
    \includegraphics[width=0.47\textwidth]{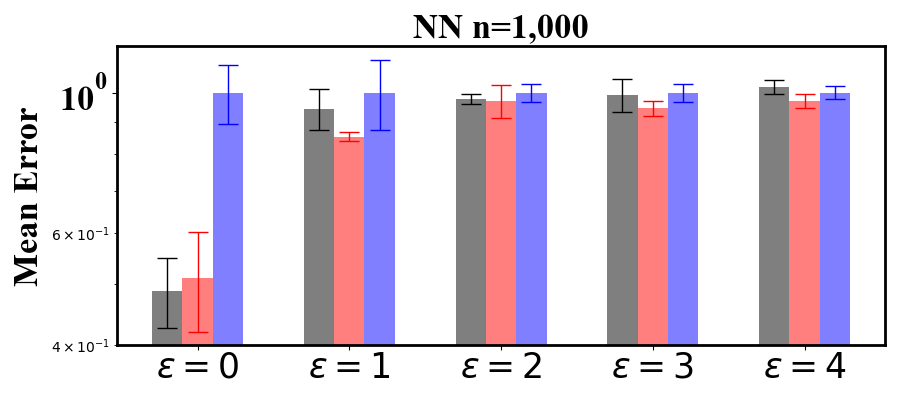}
    \includegraphics[width=0.47\textwidth]{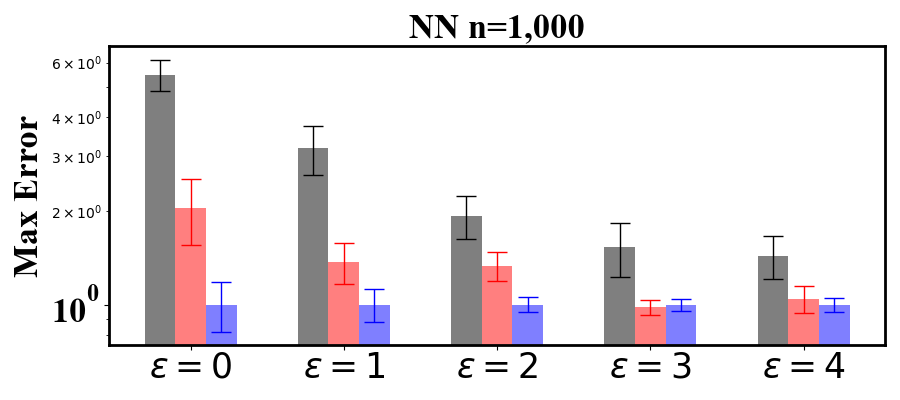}
    \includegraphics[width=0.47\textwidth]{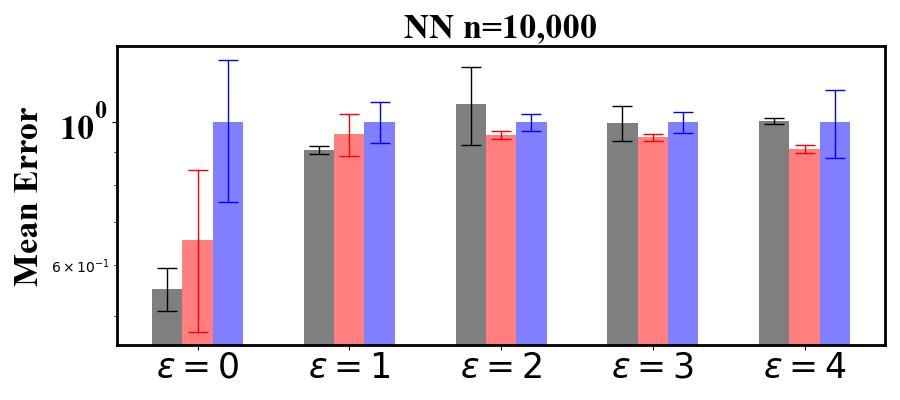}
    \includegraphics[width=0.47\textwidth]{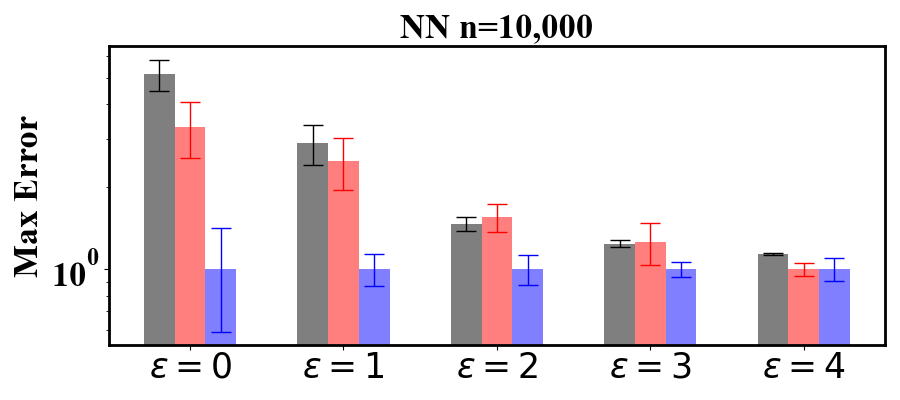}
    \caption{Statistics of the mean (left) and maximal error (right) obtained over five repetitions of downselection and training of a neural network on the synthetic dataset for random sampling (\mythickline{black}), stratified sampling (\mythickline{red}), and Algo.~\ref{algo:iterative} (\mythickline{blue}). Bar height shows ensemble mean and error bar shows standard deviation. Bar and error bar size are rescaled by the ensemble mean obtained with Algo.~\ref{algo:iterative}}
    \label{fig:NNResults}
\end{figure}

Fig.~\ref{fig:GPResults} shows the results obtained with the Gaussian process. For different levels of noise, the optimal mean absolute error is $\frac{2}{\pi} \varepsilon$, and the optimal maximal error is obtained by computing the expected value of the maximum of a random normal variable out of $N$ samples. Absolute and optimal mean and maximal error estimates are provided in the Supplementary Material. In absence of noise, Algo.~\ref{algo:iterative} ($M=10^5$, 2 iter.) offers the lowest maximal error and outperforms the other sampling methods by several orders of magnitude. In turn, the mean absolute error is also the lowest, albeit by a smaller margin than the maximal error. For the neural network case (Fig.~\ref{fig:NNResults}), the maximal error is lower by nearly an order of magnitude, but the average error is larger than the random and stratified sampling. This result also highlights how good coverage of the phase-space improves the robustness of the model. However, robustness is not well described by common metrics like mean absolute error, which is biased towards high-probability data points. 

For the neural network and the Gaussian process, Algo.~\ref{algo:iterative} exhibits the lowest maximal absolute error (close to the optimal value) for low levels of noise ($0<\varepsilon \leq 2$), but the mean absolute error is the largest out of all the methods. In particular, in the case of the Gaussian process, while the mean error was the lowest in absence of noise, it is the highest even at low noise levels. However, it should be noted that the mean error remains close to the optimal value with Algo.~\ref{algo:iterative}, with the mean error penalty relative to the other approaches being small in comparison to the unavoidable error due to noise in the data. In contrast, the other sampling approaches can result in maximal absolute errors that are significantly larger than the optimal value. These results could be explained by the reduced ability of a uniform phase-space sampling to correctly estimate the conditional mean of a noisy variable. In other words, because uniform phase-space sampling eliminates redundancy by design, it cannot benefit from redundancy when estimating a conditional average when there is nonzero conditional variance. As the level of noise increases, reducing statistical error in estimating the conditional average for high-probability regions becomes more beneficial than exploring new regions of phase-space to provide better predictions for rare points. This benefits random sampling over Algo.~\ref{algo:iterative}, leading to the observed performance trends as noise levels increase (Fig.~\ref{fig:GPResults}~and~Fig.~\ref{fig:NNResults}).

This analysis confirms that the results shown in Fig.~\ref{fig:GPNNResultsSRCPROG} can also be explained by the presence of a low level of noise for $\dot{\omega}$. Although the maximal error is lowest with Algo.~\ref{algo:iterative}, it is outperformed by other methods in terms of the mean absolute error because these methods emphasize accuracy for high-probability events and allow redundancy to counteract noise. 

\section{Conclusions}
\label{sec:conclusions}

A novel data reduction scheme was introduced in the form of an instance selection method. The method enables uniformly covering the phase-space of a dataset with an arbitrarily low number of data points. Uniform sampling is implemented by constructing an acceptance probability for each data point that depends on its probability of appearing in the full dataset. To obtain the probability map of the dataset, a normalizing flow is trained on a subsample of the full dataset. Iterative estimation of the probability map allows efficient treatment of rare events in the dataset. The advantage of the present method lies in its efficient parameterization of the probability map, which holds even for high dimensions.

The instance selection method can serve several purposes, including efficient visualization, data dissemination, and data-efficient model training. For the latter, the performance of the instance selection method was compared to other common methods. In the absence of noise, uniform phase-space sampling offers better robustness. However, the lack of redundancy of data can induce error in approximating the optimal estimator of a variable if it is noisy.

Allowing targeted redundancy in the data could be implemented in the same framework by biasing the acceptance probability of the data points towards noisy regions of the phase-space. Likewise, a refined description of the phase-space could be obtained by allocating more points in regions of large gradients. These extensions are currently being pursued and will be investigated in future publications.


\addcontentsline{toc}{section}{Appendices}
\section*{Appendix}
\renewcommand{\thesubsection}{\Alph{subsection}}

\subsection{Visual evaluation of the effect of iterations}
\label{app:visualInspection}

\begin{figure}[h!]
    \centering
    \includegraphics[width=0.31\textwidth]{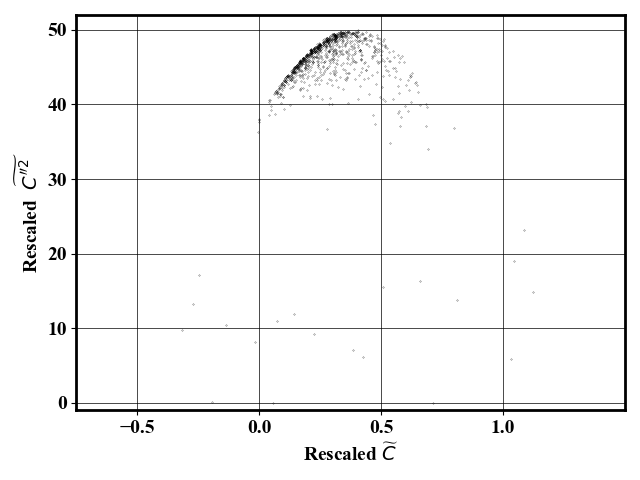}
    \includegraphics[width=0.31\textwidth]{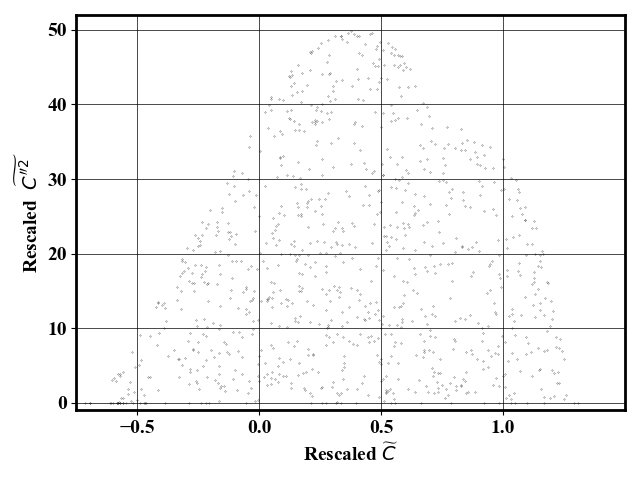}
    \includegraphics[width=0.31\textwidth]{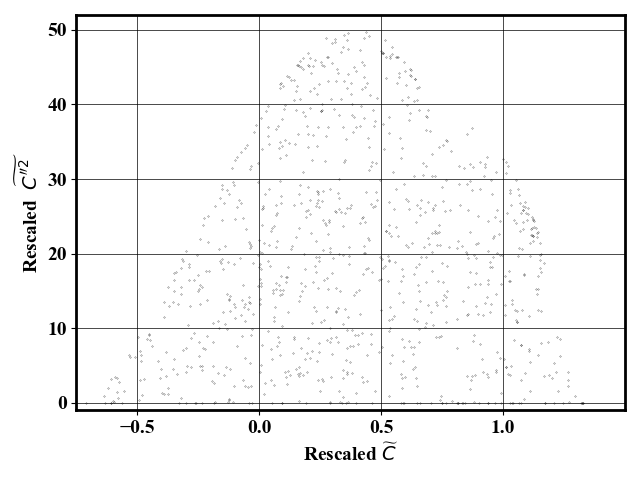}
    \includegraphics[width=0.31\textwidth]{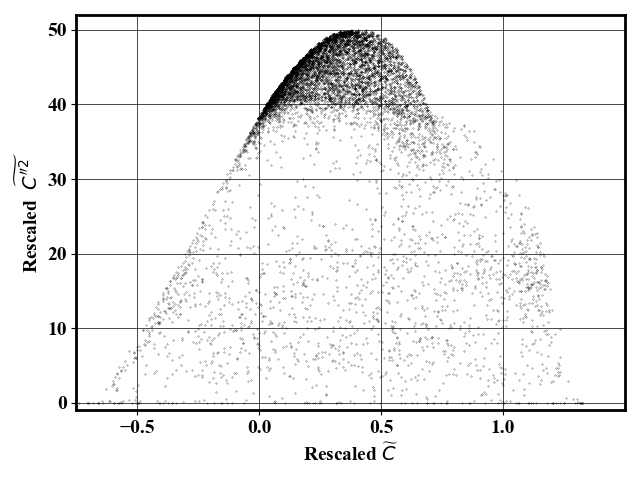}
    \includegraphics[width=0.31\textwidth]{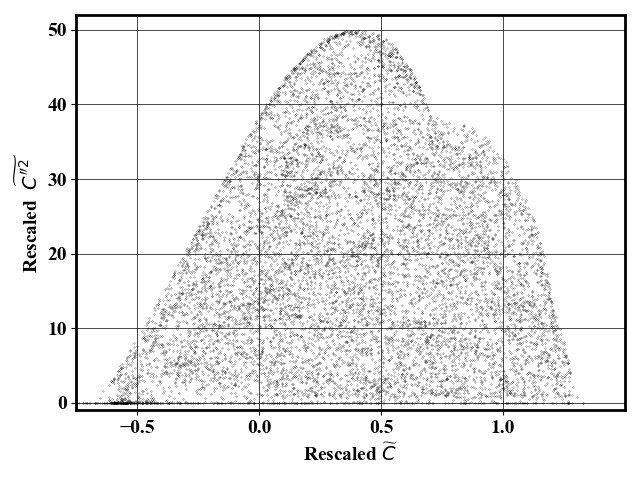}
    \includegraphics[width=0.31\textwidth]{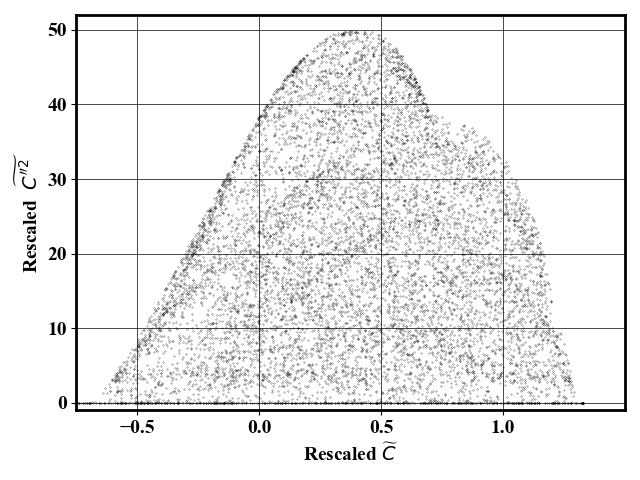}
    \caption{Illustration of the effect of using more than two iterations to select $n=1,000$ (top) and $n=10,000$ (bottom) data points with one iteration (left), two iterations (middle), and three iterations (right) of Algo.~\ref{algo:iterative}}
    \label{fig:effectIteration}
\end{figure}

The effect of using more than two iterations is illustrated in this section. As can be seen in Fig.~\ref{fig:effectIteration}, using more than two iterations does not significantly improve the uniformity of the distribution of samples. This observation holds for $1,000$ (left) and $10,000$ (right) samples. This is consistent with the results suggested by the distance criterion shown in Tab.~\ref{tab:criterionVSiteration}.

\begin{figure}[h!]
    \centering
    \includegraphics[width=0.31\textwidth]{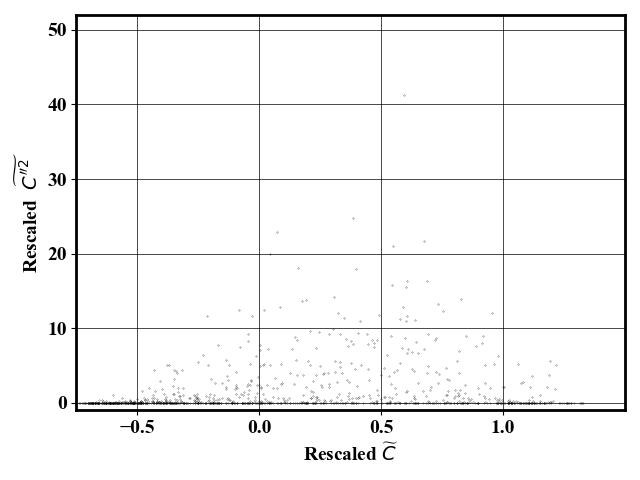}
    \includegraphics[width=0.31\textwidth]{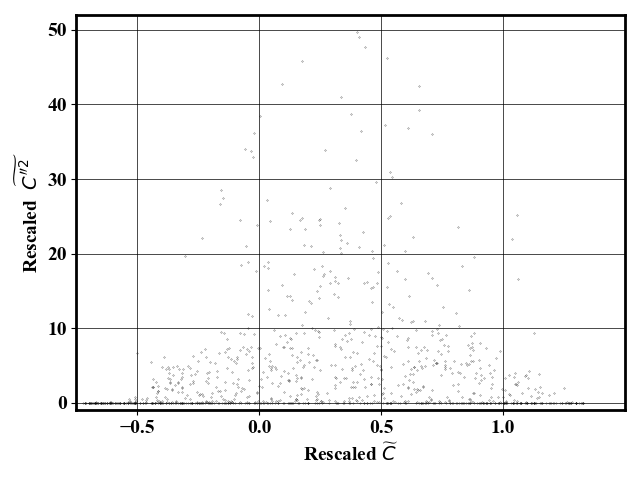}\hfill\\
    
    \includegraphics[width=0.31\textwidth]{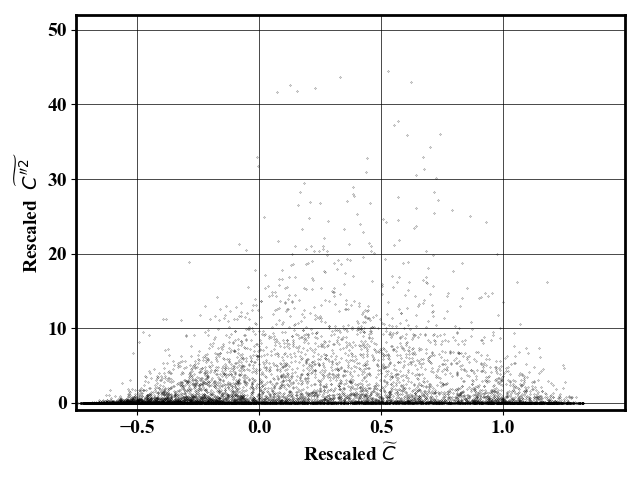}
    \includegraphics[width=0.31\textwidth]{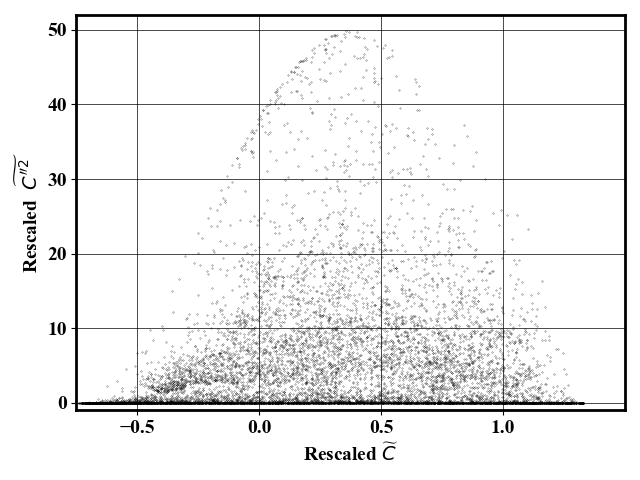}
    \caption{Illustration of random (left) and stratified (right) selection schemes of $n=1,000$ (top) and $n=10,000$ (bottom) data points}
    \label{fig:comparisonkmeansrandom}
\end{figure}

The results should be compared to random and stratified sampling for reference, shown in Fig.~\ref{fig:comparisonkmeansrandom}. It can be observed that the relative improvement obtained by using three iterations instead of two is minimal compared to random or stratified sampling.

\begin{figure}[h!]
    \centering
    \includegraphics[width=0.31\textwidth]{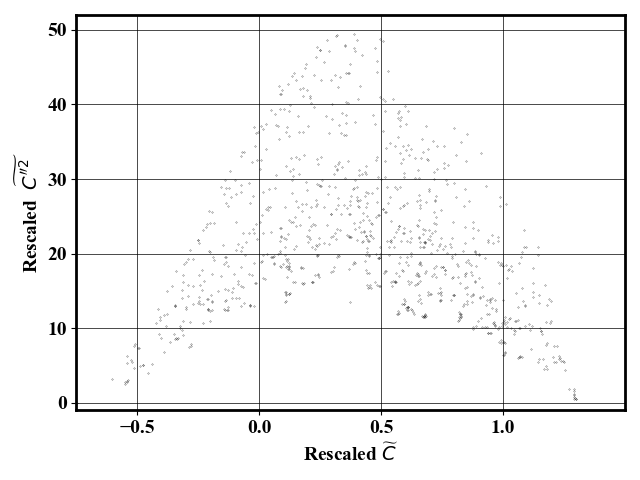}
    \includegraphics[width=0.31\textwidth]{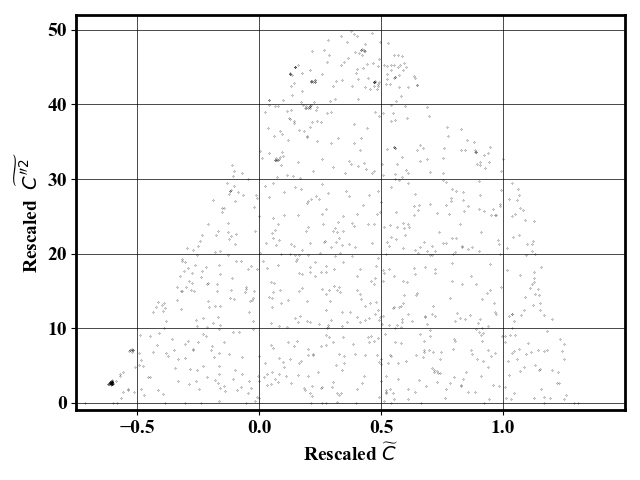}
    \includegraphics[width=0.31\textwidth]{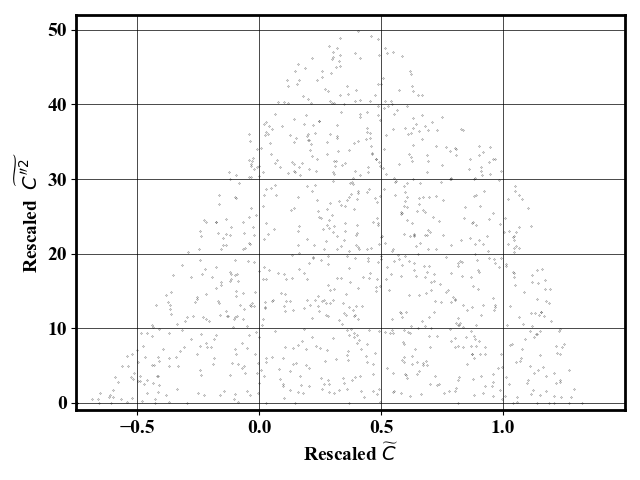}
    \includegraphics[width=0.31\textwidth]{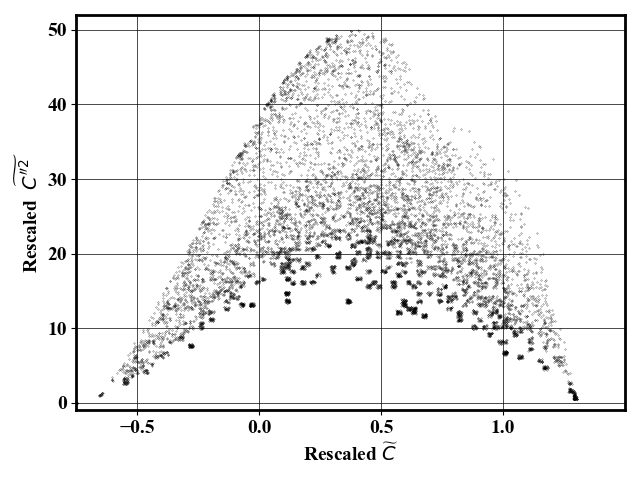}
    \includegraphics[width=0.31\textwidth]{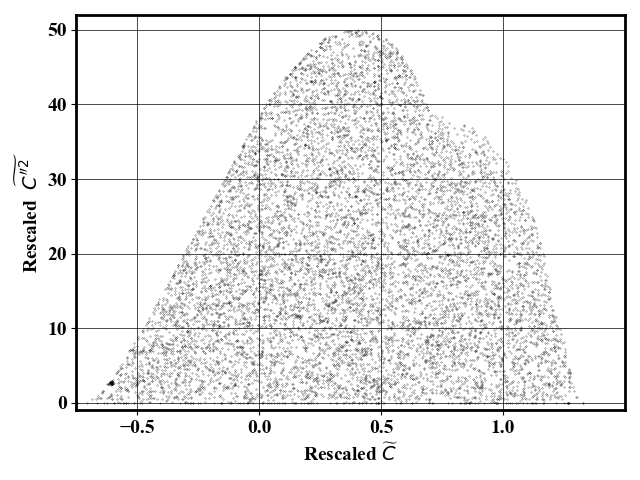}
    \includegraphics[width=0.31\textwidth]{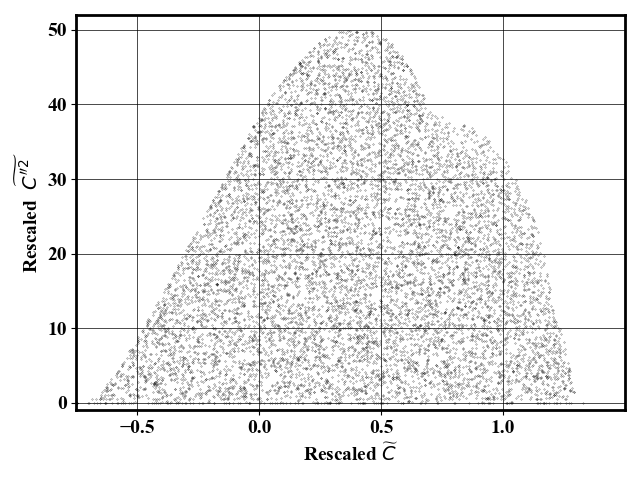}
    \caption{Illustration of the effect of using more than two iterations to select $n=1,000$ (top) and $n=10,000$ (bottom) data points with one iteration (left), two iterations (middle), and three iterations (right) of Algo.~\ref{algo:iterative}, using a binning method for the density estimation.}
    \label{fig:effectIterationBin}
\end{figure}

Using a binning strategy for the density estimation leads to results similar to using normalizing flow as shown in Fig.~\ref{fig:effectIterationBin}. While the first iteration oversamples rare data points, the next iterations lead to more uniform distribution of the downsampled data points.

\subsection{Downsampling in a different dataset}
\label{app:buildingData}

The algorithm for uniform-in-phase-space data selection is generally suitable for large $N$, moderate $D$ datasets and is not limited to the turbulent combustion and synthetic datasets described in this work. To demonstrate this, the algorithm is applied to sensor data relating to building power consumption on the campus of the National Renewable Energy Laboratory (NREL) in Golden, CO. NREL maintains instrumentation to track and record a wide array of data regarding power consumption in its buildings and weather conditions on campus to enable research into the integration of renewable energy and building efficiency technologies. Here, a dataset containing four streams of data corresponding to net power consumption in kW at NREL facilities taken at one minute intervals for the period from January 1, 2017 to November 30, 2022 is considered. The particular facilities selected were chosen to minimize the number of sampled time instances with missing data from any of the considered sensors; after discarding all time instances with missing data this results in a dataset with $N=2.7 \times 10^6$ and $D=4$. In the present use case and for energy systems in general, a data-driven model model must be available to maintain a high reliability. Therefore, a reduced dataset must contain rarely observed conditions to ensure they are accounted for. This problem is prototypical of a broader class of problems where streaming data from a limited set of sensors may be available over a long period of time and robustness to rare events is important.

\begin{figure}[h!]
    \centering
    \includegraphics[width=0.30\textwidth]{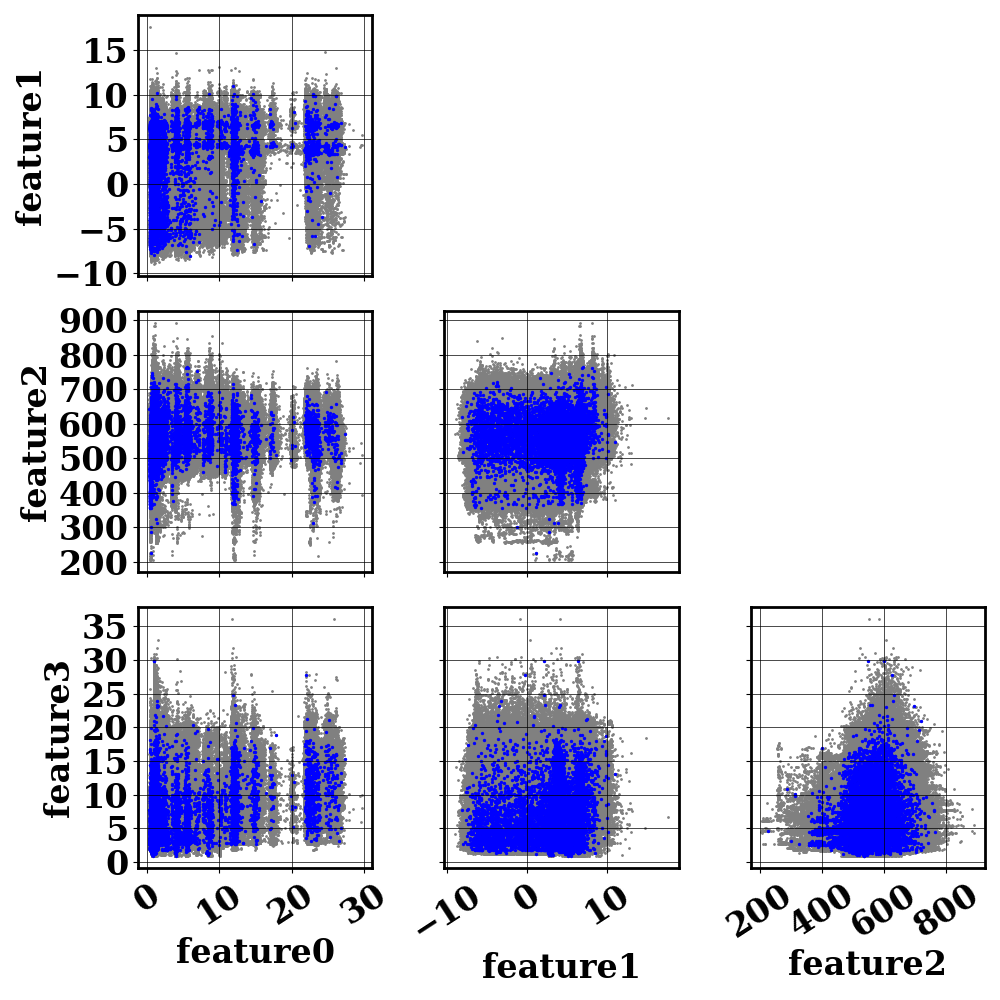}
    \includegraphics[width=0.30\textwidth]{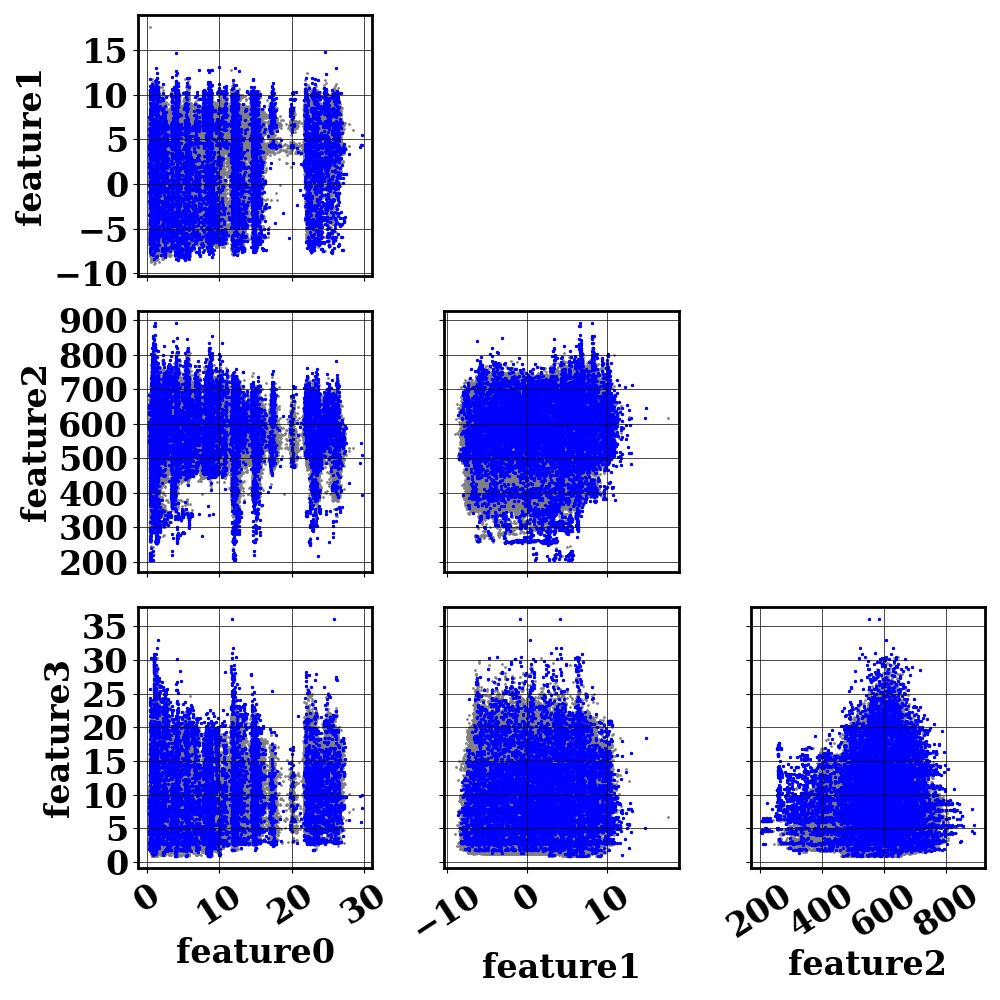}
    \includegraphics[width=0.30\textwidth]{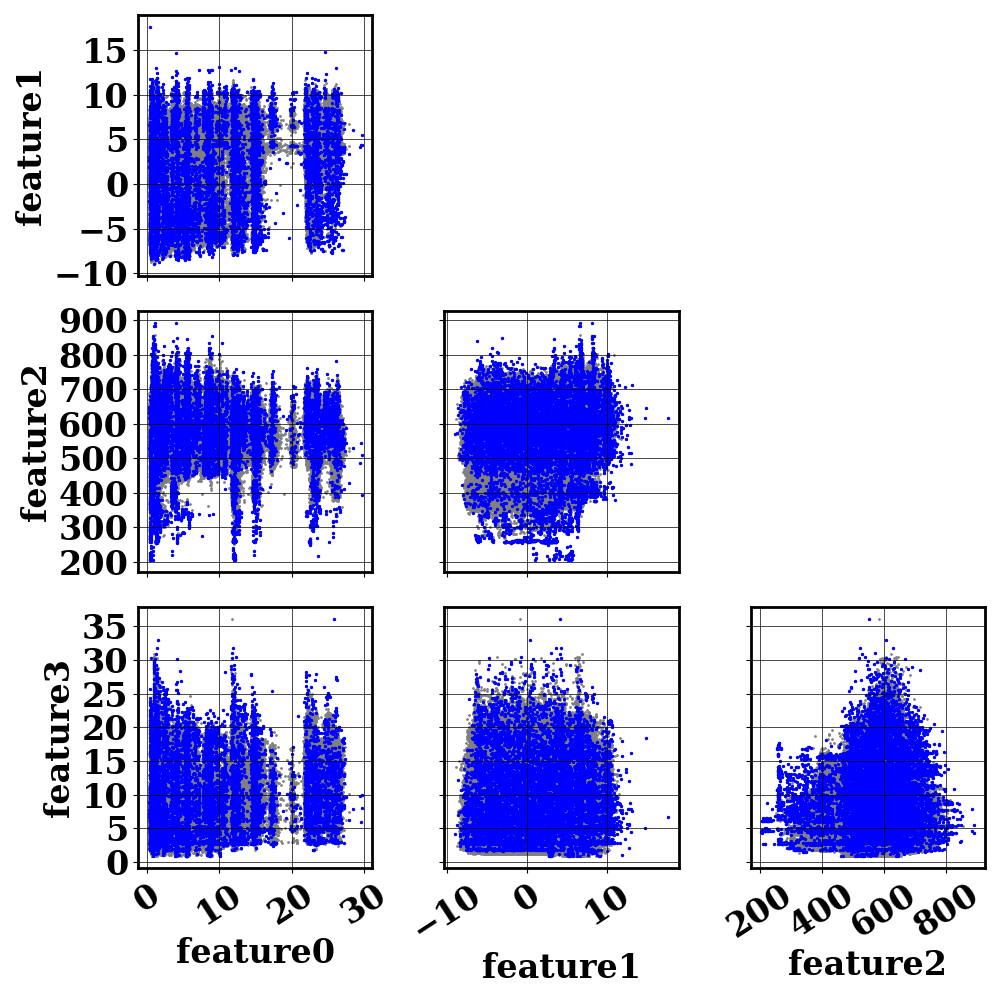}
    \caption{Two-dimensional projection of $n=10,000$ selected data points (\mydot{blue}) out of an original four-dimensional building power consumption dataset (\mydot{gray}). Left: random sampling. Middle: Algo.~\ref{algo:iterative} (2 iter.) Right: Algo.~\ref{algo:iterative} (3 iter.) }
    \label{fig:buildingData}
\end{figure}

The sampled data points using Algo.~\ref{algo:iterative} and random sampling are compared in Fig.~\ref{fig:buildingData}. As was observed for the turbulent combustion dataset in Fig.~\ref{fig:highDimCornerPlot}, Algo.~\ref{algo:iterative} leads to substantially more frequent sampling of points in the sparser regions of phase space. Notably, the character of the distribution of this dataset is very different than the previously considered dataset; in this case the density of data points is strongly multi-modal, with data concentrated in bands around particular power consumption values for each facility. Despite this multi-modal character, the iterative procedure using normalizing flows to estimate the probability distribution works similarly as before. Algo.~\ref{algo:iterative} yields similar results with 2 and 3 iterations, indicating that the conclusion that two iterations are sufficient may generalize across different datasets. This is quantitatively confirmed by the data shown in Table~\ref{tab:buildingData}. The distance criterion values are similar with two and three iterations, while both are substantially improved relative to the non-iterative Algo.~\ref{algo:base} and random sampling.

\begin{table}[h]
\caption{Dependence of the distance criterion value on the number of iterations and instances selected ($n$) for the building power consumption dataset. Larger distance criterion value is better.}
\label{tab:buildingData}
\begin{tabular}{ |c|c|c|c|c|c| } 
\hline
 Norm. flow & Algo.~\ref{algo:base} & Algo.~\ref{algo:iterative} (2 iter.) & Algo.~\ref{algo:iterative} (3 iter.)  & Random \\
\hline 
 $n=1,000$ & $0.412 \pm 0.024$ & $\boldsymbol{0.507 \pm 0.010}$ & $0.491 \pm 0.008$ & $0.280 \pm 0.008$ \\
 \hline
 $n=10,000$ & $0.225 \pm 0.004$ &  $0.247 \pm 0.001$ & $\boldsymbol{0.248 \pm 0.001}$ & $0.144 \pm 0.002$  \\
 \hline
 \end{tabular}
 \end{table}


\end{document}